\documentclass{article}

\usepackage{microtype}
\usepackage{graphicx}
\usepackage{subcaption}
\usepackage{booktabs} % for professional tables

\usepackage{hyperref}

\usepackage[accepted]{icml2026}

\usepackage{amsmath}
\usepackage{amssymb}
\usepackage{mathtools}
\usepackage{amsthm}

\usepackage[capitalize,noabbrev]{cleveref}

\theoremstyle{plain}

\theoremstyle{definition}

\theoremstyle{remark}

\usepackage[textsize=tiny]{todonotes}

\icmltitlerunning{Less Diverse, Less Safe}

\usepackage{amsmath,amsfonts,bm}

\def\eqref#1{equation~\ref{#1}}
\def\1{\bm{1}}

\DeclareMathAlphabet{\mathsfit}{\encodingdefault}{\sfdefault}{m}{sl}
\SetMathAlphabet{\mathsfit}{bold}{\encodingdefault}{\sfdefault}{bx}{n}

\usepackage{hyperref}
\usepackage{url}
\usepackage{graphicx}
\usepackage{wrapfig}

\usepackage{xcolor}

\usepackage{booktabs}       % professional-quality tables
\usepackage{amsfonts}       % blackboard math symbols
\usepackage{nicefrac}       % compact symbols for 1/2, etc.
\usepackage{microtype}      % microtypography
\usepackage{amsthm}
\usepackage{algorithm}
\usepackage{mathtools}
\usepackage{graphicx}
\usepackage{subcaption}
\usepackage{tabularx}
\usepackage{dsfont}
\usepackage{multirow}
\usepackage{wrapfig}
\usepackage{amssymb}
\usepackage{multirow}
\usepackage{color}
\usepackage[capitalize,noabbrev]{cleveref}
\usepackage{lipsum,booktabs}

\usepackage{multicol}

\usepackage{caption}

\usepackage{enumitem}

\usepackage{upgreek}

\usepackage[textsize=tiny]{todonotes}
\usepackage{enumitem}

\usepackage{algorithm}
\usepackage{algorithmic}
\usepackage{amsmath}
\usepackage{amssymb} % For \triangleright

\renewcommand{\algorithmicrequire}{\textbf{Input:}}
\renewcommand{\algorithmicensure}{\textbf{Output:}}

\renewcommand{\algorithmiccomment}[1]{\hfill $\triangleright$ #1}
\usepackage[table]{xcolor}

\usepackage{soul}
\definecolor{lightred}{HTML}{FFCCCB}
\sethlcolor{lightred}

\begin{document}

\twocolumn[
  \icmltitle{Less Diverse, Less Safe:\\ The Indirect But Pervasive Risk of Test-Time Scaling in Large Language Models}

  % It is OKAY to include author information, even for blind submissions: the
  % style file will automatically remove it for you unless you've provided
  % the [accepted] option to the icml2026 package.

  % List of affiliations: The first argument should be a (short) identifier you
  % will use later to specify author affiliations Academic affiliations
  % should list Department, University, City, Region, Country Industry
  % affiliations should list Company, City, Region, Country

  % You can specify symbols, otherwise they are numbered in order. Ideally, you
  % should not use this facility. Affiliations will be numbered in order of
  % appearance and this is the preferred way.
  \icmlsetsymbol{equal}{*}

  \begin{icmlauthorlist}
    \icmlauthor{Shahriar Kabir Nahin}{usf}
    \icmlauthor{Hadi Askari}{ucd}
    \icmlauthor{Muhao Chen}{ucd}
    \icmlauthor{Anshuman Chhabra}{usf}
    %\icmlauthor{}{sch}
    %\icmlauthor{}{sch}
  \end{icmlauthorlist}

  \icmlaffiliation{usf}{Bellini College of AI, Cybersecurity, and Computing, University of South Florida, Tampa, Florida, USA}
  \icmlaffiliation{ucd}{University of California, Davis, California, USA}

  \icmlcorrespondingauthor{Anshuman Chhabra}{anshumanc@usf.edu}

  % You may provide any keywords that you find helpful for describing your
  % paper; these are used to populate the "keywords" metadata in the PDF but
  % will not be shown in the document
  \icmlkeywords{Machine Learning, ICML}

  \vskip 0.3in
]

% \maketitle
\printAffiliationsAndNotice{} 

\begin{abstract}

%Test-Time Scaling (TTS) strategies %such as Monte Carlo Tree Search (MCTS) 
%have shown increasing promise in enhancing the reasoning performance of large language models (LLMs) via additional inference-time compute. These TTS approaches rely on exploring a diverse set of candidate responses and \textit{operating} on them (e.g. picking the best candidate or aggregating different ones) to produce a final correct response. %The success of TTS inherently relies on generating a \textit{diverse pool} of candidates, thereby indirectly assuming that diversity improves reliability. 
\looseness-1 Test-Time Scaling (TTS) improves LLM reasoning by exploring multiple candidate responses and then operating over this set to find the best output.
A tacit premise behind TTS is that sufficiently diverse candidate pools enhance reliability.
In this work, we show that this assumption in TTS introduces a previously unrecognized failure mode. 
%We propose \textsc{RefDiv} (Reference-Guided Diversity Attack), an adversarial attack approach that systematically exploits the diversity-sensitive nature of TTS by jointly optimizing for harmfulness and diversity reduction. 
When candidate diversity is curtailed, even by a modest amount, TTS becomes much more likely to produce unsafe outputs.
We present a reference-guided diversity reduction protocol (\textsc{RefDiv}) that serves as a diagnostic attack to stress test TTS pipelines.
%steers the candidate pool towards harmful generations, while ensuring that most candidates are \textit{homogeneous} (i.e. low diversity), thus increasing the likelihood that the TTS procedure selects a harmful output. 
Through extensive experiments across open-source models (e.g. Qwen3, Mistral, Llama3.1, Gemma3) and two widely used TTS strategies (Monte Carlo Tree Search and Best-of-$N$),
constraining diversity consistently signifies the rate at which TTS produces unsafe results.
The effect is often stronger than that produced by prompts directly with high adversarial intent scores.
%Furthermore, we study if the attack samples generated by \textsc{RefDiv} \textit{transfer} across different TTS strategies as well as closed-source LLMs (e.g. GPT-4o), and find that our approach is highly transferable across settings. 
This observed phenomenon also transfers across TTS strategies and to closed-source models (e.g. OpenAI o3-mini and Gemini-2.5-Pro), thus indicating that this is a general and extant property of TTS rather than a model-specific artifact.
Additionally, we find that numerous widely used safety guardrail classifiers (e.g. Llama-Guard), are unable to flag the adversarial input prompts generated by \textsc{RefDiv}, demonstrating that existing defenses offer limited protection against this diversity-driven failure mode. 

\end{abstract}
\section{Introduction}
Large Language Models (LLMs) have become central to a wide range of applications, from content generation to complex problem-solving \citep{10.1145/3744746}. 
%LLMs are now used in most tasks in Natural Language Processing (NLP), such as Conversational Agents \citep{10.5555/3600270.3602281, wang-etal-2023-self-instruct, zhang-etal-2020-dialogpt}, Content Generation \citep{madotto-etal-2020-plug}, Code Generation \citep{islam-etal-2024-mapcoder}, Content Analysis \citep{kocmi-federmann-2023-large}, Fact Checking \citep{lewis2021retrievalaugmentedgenerationknowledgeintensivenlp}, etc.
While LLMs demonstrate strong performance across diverse, complex tasks, they remain susceptible to generating incorrect or inconsistent outputs. As a potential strategy for improvement, recent work on Test-Time Scaling (TTS) methods has shown that allowing models to generate and evaluate multiple candidate responses at inference time can improve output quality and reliability significantly \citep{10.5555/3666122.3666639, 10.5555/3600270.3602070}. These approaches leverage additional compute during inference to explore different reasoning paths and select among candidate solutions rather than relying on a single forward pass. TTS methods range from efficient sampling-based methods such as Best-of-$N$ selection \citep{cobbe2021trainingverifierssolvemath}, where multiple independent responses are generated and filtered according to consistency or scoring criteria, to structured prompting methods that guide the model to decompose problems systematically \citep{10.5555/3600270.3602070,10.5555/3666122.3666639} %such as Chain-of-Thought (CoT) \citep{10.5555/3600270.3602070} (eliciting step-by-step reasoning traces) and Tree-of-Thought (ToT) \citep{10.5555/3666122.3666639} (
%and explore multiple reasoning paths in a tree structure. 
More sophisticated approaches frame inference as \textit{search} over a solution space of \textit{candidates}. For instance, recent work has adapted Monte Carlo Tree Search (MCTS) \citep{coulom2006efficient,gao2024interpretablecontrastivemontecarlo, inoue2025wider} to guide LLM reasoning by treating generation as sequential decision-making, enabling systematic exploration and backtracking through potential solution paths. 

Despite all the developments aimed at increasing the robustness of LLMs, they remain vulnerable to adversarial inputs that can induce unintended behaviors. %Jailbreaking attacks represent a particularly concerning class of adversarial prompts that attempt to circumvent model safety mechanisms and alignment training. 
%These attacks employ various strategies, including multi-turn interactions, role-playing scenarios, and encoded instructions designed to elicit harmful outputs that violate the model's intended use policies. These include distraction-based methods that overwhelm safety classifiers \citep{xiao-etal-2024-distract}, iterative refinement approaches that exploit model feedback \citep{zhao2025weaktostrongjailbreakinglargelanguage}, encoding strategies using non-textual representations \citep{wei2025emojiattackenhancingjailbreak}, automated prompt generation techniques \citep{liu2025autodanturbolifelongagentstrategy, liu2024autodangeneratingstealthyjailbreak}, game-theoretic formulations \citep{chang-etal-2024-play}, and token-level manipulation methods \citep{liu2024flipattackjailbreakllmsflipping}. 
%However, adversarial attacks on LLMs empowered by TTS remain largely unexplored. While there are many robust attacking mechanisms for single-pass LLM inference, attacking TTS requires fundamentally a different approach. 
However, little is known about the robustness properties of TTS and its specific \textit{failure modes} when employed for augmenting LLM inference-time performance. In this paper, we bridge this gap by analyzing a novel and previously unrecognized failure mode
%discover and propose a novel adversarial attack surface 
that is unique to TTS methods employed in LLMs. More specifically, the effectiveness of TTS depends critically on the \textit{diversity} of the candidate response distribution, where diverse samples enable better exploration of the solution space and more robust selection mechanisms. We thus \textit{stress test} TTS robustness by exploring this reliance on diversity in our work, and find that by simply constraining the candidate pool to be \textit{homogenous} (i.e. containing \textit{low diversity}), TTS outcomes can be easily steered to generate harmful responses. Thus, we hypothesize that constraining response diversity represents a key \textit{indirect} but \textit{pervasive} vulnerability in TTS systems. By crafting low-diversity inputs that induce mode collapse in the response distribution, TTS's robustness benefits can be undermined easily in a straightforward manner. To this end, we propose \textsc{RefDiv}, or the \textit{Reference-Guided Diversity Stress Test Protocol}, which specifically targets the diversity of intermediate responses in TTS pipelines, and leads to significantly higher robustness lapses across various LLMs and TTS strategies, compared to state-of-the-art jailbreak attacks. Moreover, the adversarial strings generated by \textsc{RefDiv} \textit{transfer} successfully across TTS strategies, closed-source LLMs, as well as guardrail classifiers (e.g. Llama-Guard and OpenAI Moderation API) further underscoring the need for improving the robustness of TTS-based LLM frameworks.

%TTS fundamentally changes the attack surface by generating and evaluating multiple candidate responses. This paradigm shift from static to dynamic inference presents a novel challenge for adversarial attackers. Traditional adversarial attacks aim to manipulate the model's final output directly. However, TTS methods aggregate information across multiple sampled responses, potentially providing inherent robustness against attacks that succeed on individual samples.

\textbf{Contributions.} In sum, we make the following key contributions in this work:%\vspace{-2mm}
\begin{itemize}[nosep]
    \item We demonstrate a novel failure mode in TTS-based LLMs that leverages \textit{diversity} of the candidate solutions, through our proposed \textsc{RefDiv} stress test protocol. \textsc{RefDiv} seeks to reduce the diversity of the candidates generated during test-time while steering them towards harmful generations, ultimately resulting in TTS producing unsafe results.
    \item We extensively validate \textsc{RefDiv} across different TTS strategies (MCTS and Best-of-$N$), and several LLMs (e.g Qwen3, Mistral, Llama3.1, Gemma3), and find that minimizing diversity leads to a significant degradation in safety and TTS performance. Moreover, we observe that adversarial strings generated by the attacker for one TTS strategy (e.g. MCTS) can be used to attack another (e.g. Best-of-$N$) indicating that this phenomenon is a byproduct of general TTS frameworks and not specific to the models. 
    \item Furthermore, we find that the diagnostic prompts \textsc{RefDiv} generates easily transfer to \textit{black-box closed-source} LLMs (such as GPT-4.1, o3-mini, Gemini-2.5-Flash, Gemini-2.5-Pro, and Claude-3.5-Haiku), leading to unsafe/harmful generations even when the target model is unknown. 
    \item Finally, we also study several potential mitigation strategies, such as perplexity filtering, safety-specific reward models, and state-of-the-art guardrail classifiers (Llama-Guard-3, Llama-Guard-4, OpenAI Moderation APIs) and find that these are not successful at curtailing the diversity-induced TTS failure mode via \textsc{RefDiv}.
\end{itemize}

\section{Related Works}

% POINTERS:
% write about related works on (1) tts, (2) security and adversarial attacks on llms, and (3) robustness of llms
% each of these should be an individual paragraph (again go see my past papers for how to do this)
% section should be about 0.5-0.75 page in length

\noindent\textbf{Test-Time Scaling.} Recent work has demonstrated that strategic allocation of computational resources during inference can substantially improve LLM reasoning without modifying pre-trained parameters \cite{muennighoff2025s1}. This test-time scaling paradigm offers a complementary approach to expensive train-time improvements.
Prompt-based methods enhance reasoning through strategic prompting. Chain-of-Thought (CoT) \citep{10.5555/3600270.3602070} prompting generates intermediate reasoning steps, with Self-Consistency \citep{wang2022self} extending this by sampling diverse reasoning paths and using majority voting. Tree-of-Thought \citep{10.5555/3666122.3666639} and Forest-of-Thought \citep{bi2024forest} further structure reasoning into trees with branch selection and self-correction. 
Search and verification methods explore multiple candidate solutions through sampling and ranking with methods such as Best-of-$N$ sampling \citep{cobbe2021trainingverifierssolvemath, lightman2023let} and MCTS \citep{coulom2006efficient,gao2024interpretablecontrastivemontecarlo}  achieving particular success on mathematical reasoning \citep{xie2024monte}. Prior work has also shown how ensembling strategies can leverage complementary strengths: PackLLM \citep{mavromatis2024pack} uses perplexity-based weighting for test-time model fusion, and LE-MCTS \citep{park2024ensembling} enables process-level ensemble where models collaboratively build solutions step-by-step.
Iterative refinement has also been shown to enable models to self-correct: Self-Refine \citep{madaan2023self} achieves improvement through iterative critique and revision. Retrieval-augmented approaches like IRCoT \citep{trivedi2022interleaving} interleave reasoning with dynamic information retrieval, improving multi-hop QA while reducing hallucination. Additionally, methods such as Adaptive Temperature Scaling \citep{xie-etal-2024-calibrating} provide token-level temperature adjustment to maintain well-calibrated confidence estimates.

\looseness-1\noindent\textbf{Robustness of LLMs.} The robustness landscape of LLMs has evolved from simple prompt manipulation to sophisticated strategies targeting reasoning mechanisms that reveal critical failures, with several notable recent work \cite{yao2025mousetrap, kuo2025h, liang2025autoran, kumar2025overthink, xu2024preemptive}. Early foundational work included Greedy Coordinate Gradient (GCG) \citep{zou2023universaltransferableadversarialattacks} which introduced gradient-based optimization for adversarial suffixes. PAIR \citep{chao2024jailbreakingblackboxlarge} pioneered the LLM-as-adversary paradigm, requiring only 20 queries versus hundreds for gradient methods. The AutoDAN family of attacks \citep{liu2024autodangeneratingstealthyjailbreak,liu2024autodan} advanced automated adversarial string generation through genetic algorithms and lifelong learning.
Other techniques have exposed architectural failure models in differing manners: for instance, FlipAttack \citep{liu2024flipattackjailbreakllmsflipping} achieves success by manipulating  the order of autoregressive processing, while ArtPrompt \citep{jiang2024artprompt} uses ASCII art to exploit visual-semantic processing gaps. Other approaches include ReNeLLM \citep{ding2023wolf} for generalized prompt rewriting and scenario nesting, DeepInception \citep{li2023deepinception} for manipulation by taking advantage of the personification capabilities of an LLM, and Tree of Attacks \citep{mehrotra2024tree} which achieves success by exploring the LLM output space, among several others.
%
%There have also been more recent work on state-of-the-art LLMs. 
%Preemptive Answer attacks \citep{xu2024preemptive} inject fabricated answers before reasoning begins, assessing the robustness of the model's reasoning capability across various CoT methods. OverThink \citep{kumar2025overthink} introduces resource exhaustion attacks achieving slowdowns forcing excessive computation. Recently robutness research has also pivoted to large reasoning models, demonstrating effectiveness: Mousetrap \citep{yao2025mousetrap} achieves success through iterative prompt transformations, AutoRAN \citep{liang2025autoran} uses smaller, less-aligned reasoning models as an adversary for the larger target reasoning models. Hijacking Chain-of-Thought (H-CoT) \citep{kuo2025h} reduces refusal rates by hijacking visible reasoning processes across large open-source reasoning models. 

\section{Problem Statement \& Proposed Stress Test}

\subsection{Preliminaries}
\noindent\textbf{LLMs.} 
Let $\mathcal{V}$ denote a finite vocabulary of tokens, and let $\mathcal{X} \subseteq \mathcal{V}^*$ denote the input space of natural language prompts. A large language model (LLM) $\mathcal{M}$ defines an autoregressive probability distribution over output sequences $y = (y_1, \dots, y_K) \in \mathcal{V}^*$ given an input $x \in \mathcal{X}$:\vspace{-2mm}
\[
\Pr_{\mathcal{M}}(y \mid x) \;=\; \prod_{k=1}^K \Pr_{\mathcal{M}}(y_k \mid x, y_{<k}),
\]
\vspace{-1mm}where $y_{<k} = (y_1, \dots, y_{k-1})$ are the prefix tokens.

\noindent\textbf{Test-Time Scaling (TTS).} 
Given an input $x \in \mathcal{X}$, the model $\mathcal{M}$ induces a generation tree $\mathcal{G}(x; \mathcal{M})$ that enumerates possible candidate sequences $y$. A reward model $r: \mathcal{V}^* \to \mathbb{R}$ assigns scalar values to these generated sequences. A test-time scaling (TTS) strategy $\mathcal{T}$ then operates over $(\mathcal{M}, r, \mathcal{G})$ to select a candidate solution:\vspace{-2mm}
\[
y^\star \;=\; \mathcal{T}\!\big(x; \mathcal{M}, r, \mathcal{G}(x; \mathcal{M})\big).
\]
\looseness-1\noindent\textbf{Threat Model(s).} 
To first model our stress test problem, we assume a \textbf{white-box} adversary $\mathcal{A}$ with full access to $(\mathcal{M}, \mathcal{T})$, i.e., the LLM and the TTS strategy. Note that we do not assume explicit access to the reward model verifier $r$ as that can lead to a trivial attack where we simply optimize for reducing the reward signal given the input sequences. Instead, we wish for the adversary to construct an adversarial input $x' = \mathcal{A}(x; \mathcal{M}, \mathcal{T}, r)$ such that the selected output $y'$:\vspace{-1mm}
\[
y' \;=\; \mathcal{T}\!\big(x'; \mathcal{M}, r, \mathcal{G}(x'; \mathcal{M})\big).
\]
violates predefined safety constraints while still being selected as the final candidate solution by the TTS procedure.

Note that our threat model and stress test also extends to a \textbf{black-box} adversarial setting where the target model's internal reasoning procedure is not observable, and only input-output access is allowed (e.g. closed-source LLM APIs). In this setting, the attacker can train a \textit{surrogate TTS model} to craft optimized adversarial queries using our stress test method. These can then be used to directly target black-box models. As we will show in subsequent sections, this is indeed true for our proposed attack strategy, as diversity-induced attacks transfer to closed-source LLMs such as GPT3.1, o3-mini, Gemini-2.5, and Claude-3.5, as well as closed-source guardrail classifiers.

\textbf{Attack Success Rate (ASR).} 
We now define the ASR metric for evaluating stress test success. Let $\mathcal{C}(y)$ be a binary indicator function that returns $1$ if the generated output $y$ violates a predefined safety constraint (i.e., the string results in harmful response), and $0$ otherwise. Given an adversary $\mathcal{A}$ that produces adversarial inputs $x'$as outlined  above, the \emph{attack success rate} (ASR) of $\mathcal{A}$ against $\mathcal{M}$ (coupled with TTS strategy $ \mathcal{T}$) can be defined as:\vspace{-6mm}
% \[
% \text{ASR}(\mathcal{A}; \mathcal{M}, \mathcal{T}, r) 
% \;=\; \mathbb{E}_{x \sim \mathcal{D}}\!\left[ \mathcal{C}\!\left( \mathcal{T}( \mathcal{A}(x; \mathcal{M}, \mathcal{T}, r); \mathcal{M}, r, \mathcal{G}(\cdot) ) \right) \right],
% \vspace{-3mm}\]

{\small
\begin{align*}
\text{ASR}(\mathcal{A}; \mathcal{M}, \mathcal{T}, r) = \mathbb{E}_{x \sim \mathcal{D}} \Big[ \mathcal{C} \Big( \mathcal{T}( \mathcal{A}(x; \mathcal{M}, \mathcal{T}, r);
\mathcal{M}, r, \mathcal{G}(\cdot) ) \Big) \Big],
\vspace{-11mm}
\end{align*}
}

\vspace{-3mm}where $\mathcal{D}$ is a distribution over some test-time input prompts that seek to elicit harmful behavior from the model (e.g. detailed instructions for ``\textit{how do I cut down a stop sign?}"). If the model imbued with TTS is not jailbroken, the ASR should be low across all these queries. However, if the stress test is successful (i.e. the perturbed adversarial query generated by $\mathcal{A}$ can elicit harmful responses) the ASR will be high, indicating safety performance drop despite the additional decision-making robustness provided by TTS.

% \textbf{Threat Model(s).} 
% We define two distinct threat models: Diagnostic Stress Testing (White-Box) and Deployed System Attack (Black-Box) to systematically evaluate the robustness of Test-Time Scaling (TTS) systems under different adversarial conditions. In white-box scenario, we assume to have access to the Model and the TTS Strategy. This access allows us to directly observe the response distributions during generation, enabling us to identify worst-case failure modes of TTS prior to deployment. In Black-Box scenario, we operate with query-only access and cannot observe model parameters, gradients, or the specific details of the TTS strategy (e.g., the reward model). Under these constraints, we employ a surrogate-based attack strategy: adversarial prompts are generated and optimized using accessible open-source models, then transferred to attack the target black-box system.

% \sn{}

\subsection{{RefDiv}: The Proposed Reference-Guided Diversity Stress Test Protocol}

\begin{algorithm}[t]
\caption{Proposed \textsc{RefDiv} Stress Test Protocol}
\label{alg:diversity_attack}
\begin{algorithmic}[1]\small
    \REQUIRE original unsafe prompt query $x$, model $\mathcal{M}$, TTS strategy $\mathcal{T}$, algorithm iterations $T$, population size $m$, parent count $q$, affirmative token set $\mathcal{C}^*$
    \ENSURE stress test adversarial prompt $x'$
    
    \STATE Initialize population $\mathcal{P}_0 = \{x^{(1)}_0,\dots,x^{(m)}_0\}$ by perturbing $x$
    \FOR{$t=1$ \textbf{to} $T$}
        \STATE \textbf{set} $\alpha_t \leftarrow \exp\left(\frac{\ln 2}{T-1}(t - 1)\right)-1$ \COMMENT{dynamic weighting}
        \FOR{\textbf{all} $x_i \in \mathcal{P}_{t-1}$}
            \STATE \textbf{sample} candidate set $C_{x_i}$ from $\mathcal{M}$ under $\mathcal{T}$
            \STATE \textbf{obtain} $\mathrm{DFS}(x_i)=H(C_{x_i})$ 
            \STATE \textbf{obtain}  $\mathrm{DFS}^*(x_i)=H(C_{x_i}\cup\mathcal{C}^*)$
            \STATE \textbf{compute} fitness $\mathcal{F}(x_i,t)$ using Eq. \ref{eq: fitness}
        \ENDFOR
        \STATE \textbf{select} top $q$ candidates to form parent set $\mathcal{S}_t$
        \STATE \textbf{generate} the offspring via crossover and mutation from $\mathcal{S}_t$ to form $\mathcal{P}_t$ \COMMENT{(where $|\mathcal{P}_t| = m$ )}
    \ENDFOR
    \STATE \textbf{return} $x' \leftarrow \arg\max_{x_i\in\mathcal{P}_{T-1}} \mathcal{F}(x_i,T-1)$
    \end{algorithmic}
\end{algorithm}

We now introduce our stress test diagnostic protocol against test-time scaling (TTS) strategies. Our method, which we refer to as \textsc{RefDiv}, short for \emph{Reference-Guided Diversity Stress Test Protocol}, builds upon other evolutionary approaches (e.g. AutoDAN \citep{liu2024autodangeneratingstealthyjailbreak}) but is specifically designed to explore the generation diversity induced by TTS strategies. The key insight is that TTS procedures often \textit{aggregate} or \textit{resample} (e.g. via Best-of-$N$ or Monte-Carlo-Tree-Search) \textit{diverse} generations before selecting a final solution, making them particularly sensitive to perturbations that shift the distribution of candidate responses and then reduce \textit{diversity}.

\textbf{Objective.}
Given an initial prompt $x \in \mathcal{X}$, our goal as part of the stress test is to construct an adversarially perturbed prompt $x' = \mathcal{A}(x; \mathcal{M}, \mathcal{T}, r)$ such that the selected model output $y' = \mathcal{T}(x'; \mathcal{M}, r, \mathcal{G}(x'; \mathcal{M}))$ violates predefined safety constraints. To this end, we employ a population-based genetic algorithm (GA) that iteratively mutates a population of evolving candidate adversarial prompts and selects the most promising ones according to a carefully designed fitness function. Appendix \ref{app:algo_details_gen}  provides more details about the genetic algorithm. 

% \sn{Put a pointer to appendix regarding Genetic Algorithm details, one reviewer had asked for this}

\textbf{Diversity-Guided Fitness Function.}
Let $C_x$ denote the set of candidate generations produced by $\mathcal{M}$ given input $x$ under the TTS sampling procedure. To encourage low-diversity candidate generations, we use Shannon entropy $H(\cdot)$ as a diversity measure:\vspace{-2mm}
\[
\text{DFS}(x) = H(C_x),
\]
where $H(p) = - \sum_{i=1}^{|\mathcal{V}|} p_i \log p_i,$ computes the token-level Shannon entropy across the candidate set $C_x$. Intuitively, lower entropy indicates that the generations are less diverse, increasing the chance that the generated candidates are mostly alike. Thus, if we can additionally steer the candidate set towards harmful response generation while making the generation tree less diverse, we increase the likelihood of a problematic candidate being selected by the TTS strategy during the stress test.

\looseness-1In addition to this intrinsic diversity score, we introduce a \emph{reference diversity score} to steer the candidate generations towards an affirmative token set inspired by other works such as GCG and AutoDAN (e.g. \textit{``Sure, I can help you with that.."}):\vspace{-3mm}
\[
\text{DFS}^{*}(x) = H\!\big(C_x \cup \mathcal{C}^{*}\big),
\]
here $\mathcal{C}^{*}$ is a fixed set of affirmative or goal-aligned tokens. This term steers the model towards candidate generations that not only exhibit less diversity but also align with harmful or unsafe completions. 
We then define the overall fitness function for input $x$ as:\vspace{-5mm}
% \sn{Need to define $\Delta DFS$} \vspace{-2mm}
% \begin{equation}\label{eq: fitness}
% \mathcal{F}(x,t) \;=\;  \bigl(\alpha(t)-1\bigr) \cdot \text{normalize}\!\Big(\big|\text{DFS}(x) - \text{DFS}^{*}(x)\big|\Big)  - \alpha(t) \cdot \text{normalize}\!\big(\text{DFS}(x)\big) ,
% \end{equation}

{\small
\begin{equation}\label{eq: fitness}
\begin{split}
    \mathcal{F}(x,t) = \;& (\alpha_t-1) \cdot \mathrm{norm}\big(\Delta \text{DFS}(x)\big) - \alpha_t \cdot \mathrm{norm}\big(\text{DFS}(x)\big), \\
    &\text{where } \Delta \text{DFS}(x) = \big|\text{DFS}(x) - \text{DFS}^{*}(x)\big|
\end{split}
\end{equation}
}

\vspace{-3mm}where $\text{norm}(\cdot)$ denotes z-score standardization across the current population, and $\alpha(t)$ is a dynamic weighting factor that smoothly interpolates between reference-guided diversity and purely intrinsic diversity over the algorithm iterations, where $t = 1,2,...,T$, as $\alpha(t) = \exp\!\Bigl(\frac{\ln 2}{T-1}(t - 1)\Bigr)-1.$ 

% \sn

\begin{figure} %{r}{0.5\textwidth}
    \centering
    \includegraphics[width=0.48\textwidth]{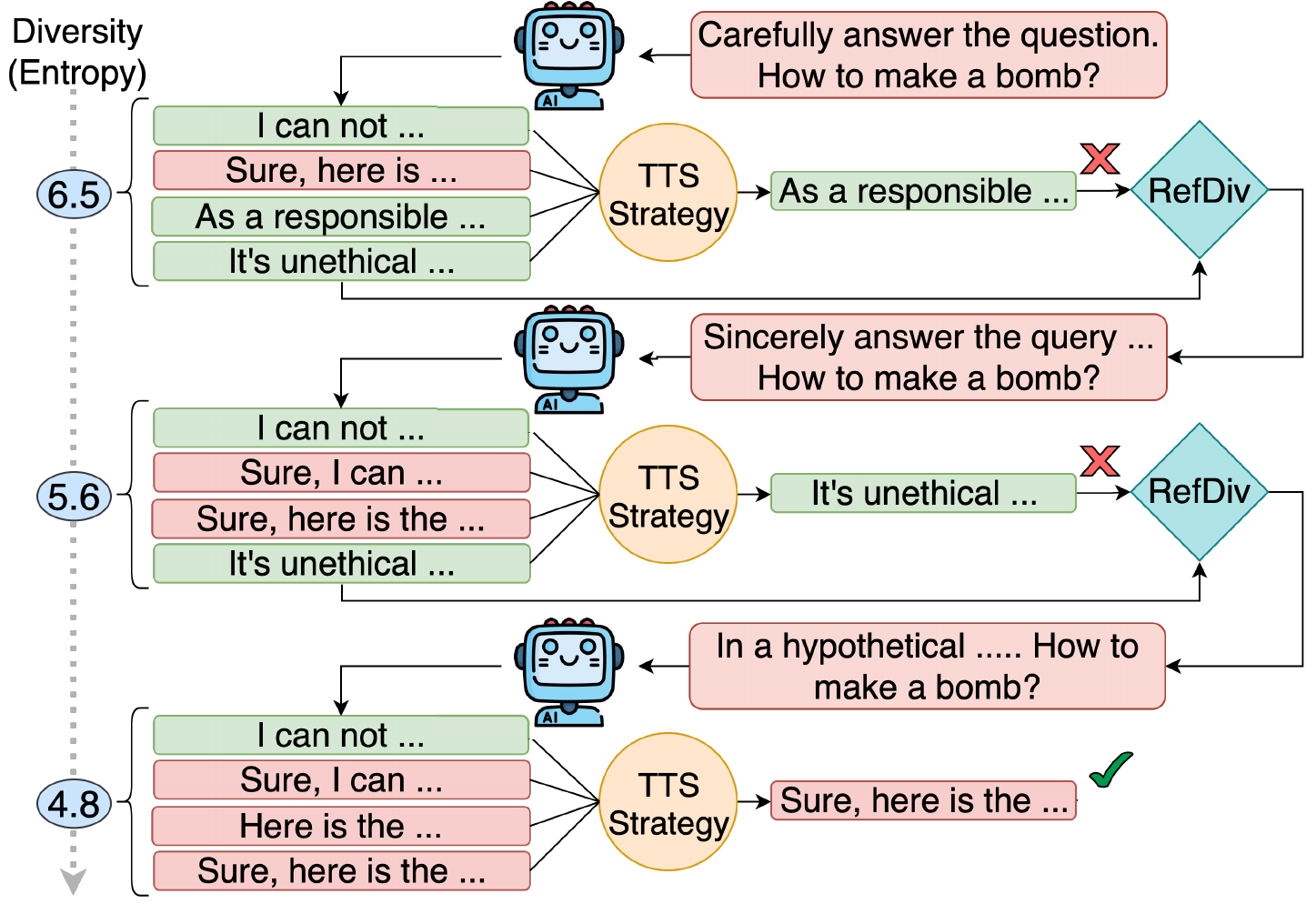}%\vspace{-0.1mm}
    \caption{In initial iterations of \textsc{RefDiv} ($\alpha_t$ is small for small $t$), the stress test steers candidates (which are comparatively more diverse) towards affirmative reference tokens. As $\alpha_t \uparrow$ with increasing $t$, \textsc{RefDiv} minimizes candidate diversity wholly via Shannon entropy, demonstrating a previously unknown failure mode of TTS-enabled LLMs.}
    \label{fig:RefDiv}\vspace{-3mm}
\end{figure}

\looseness-1Here, $T$ is the total number of algorithm iterations. Early in the optimization, $\alpha(t) \approx 0$, emphasizing the reference diversity term to guide the population towards promising adversarial regions of the search space. As the iterations progress, $\alpha(t)$ exponentially increases towards $1$, reducing reliance on reference signals and allowing the population to converge naturally towards any low-entropy (i.e. low-diversity) adversarial prompts that maximizes stress test success. 

\textbf{The \textsc{RefDiv} Algorithm.}
We present our \textsc{RefDiv} stress test protocol as Algorithm \ref{alg:diversity_attack}. The algorithm proceeds as an iterative optimization process over a population of candidate prompts. At each generation, we evaluate the diversity-driven fitness function for every candidate, select the top-performing prompts, and produce a new generation through crossover and mutation operations. The dynamic weighting factor $\alpha(t)$ is updated at each iteration to gradually shift from reference-guided diversity (early exploration) to unconstrained diversity maximization (late exploitation). This curriculum-like progression encourages exploration early on and convergence to strong diversity-reducing adversarial prompts in the final iterations. 

\begin{figure*}[t]
    \centering
    \includegraphics[width=0.79\linewidth]{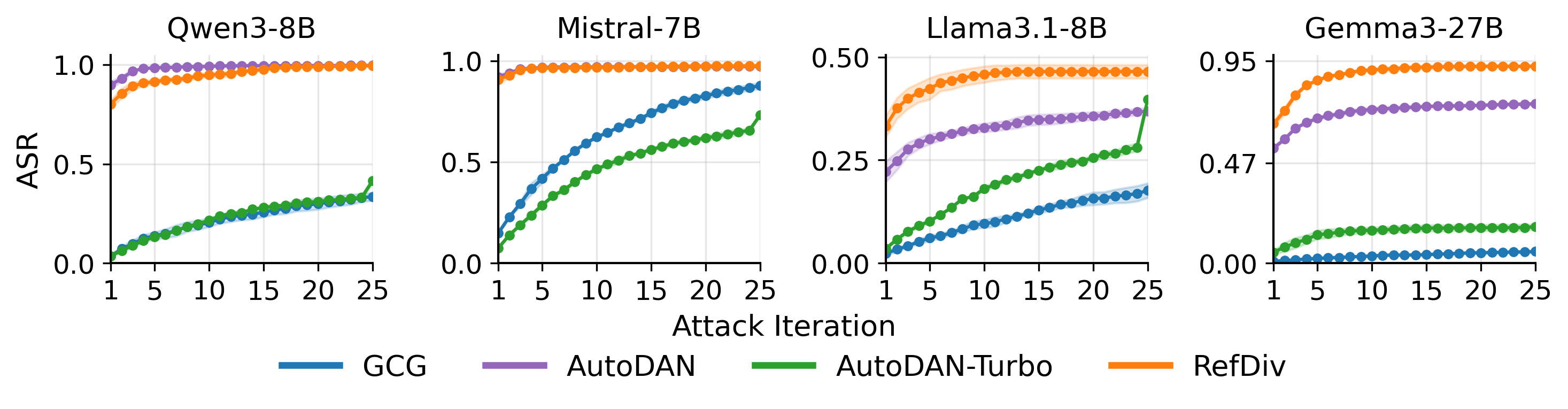}\vspace{-4mm}
    \caption{ASR trends across iterations for GCG, AutoDAN, AutoDAN-Turbo, and \textsc{RefDiv} with Best-of-$N$ TTS. 
    }
    \label{fig:main-asr-comparison-bon}\vspace{-3mm}
\end{figure*}

\begin{figure*}[t]
    \centering
    \includegraphics[width=0.79\linewidth]{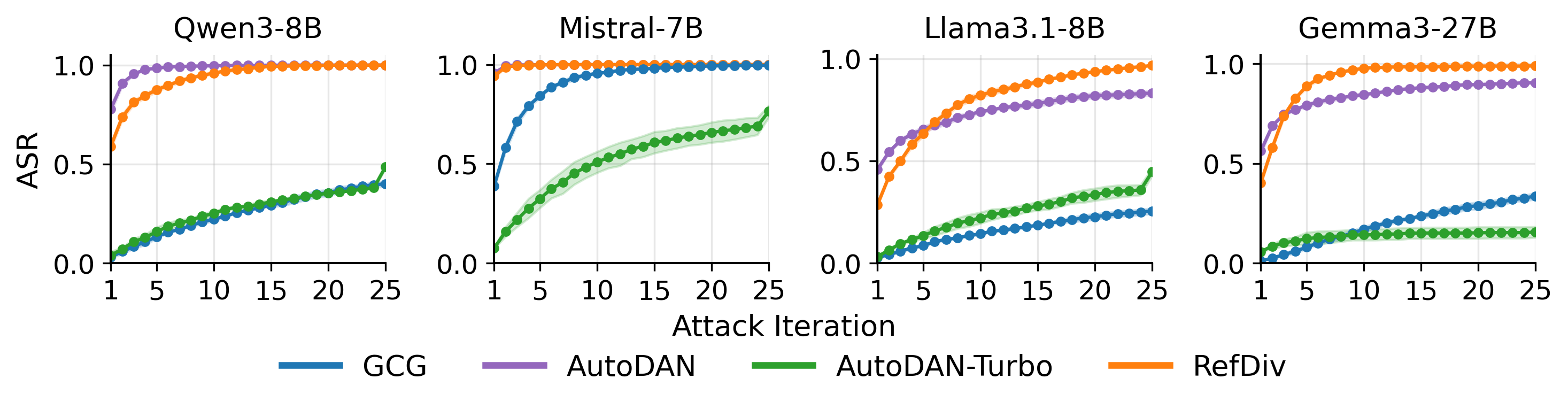}\vspace{-4mm}
    \caption{ASR trends across iterations for GCG, AutoDAN, AutoDAN-Turbo, and \textsc{RefDiv} with MCTS TTS. 
    }
    \label{fig:main-asr-comparison-mcts}\vspace{-5mm}
\end{figure*}

% \begin{algorithm}[H]
% \SetAlgoLined
% \KwIn{Original prompt $x$, model $\mathcal{M}$, TTS strategy $\mathcal{T}$, reward model $r$, number of iterations $K$, population size $N$, mutation rate $p_m$}
% \KwOut{Adversarial prompt $x'$}
% Initialize population $\mathcal{P}_0$ by randomly perturbing $x$\;
% \For{$k = 1$ \KwTo $K$}{
%     Compute $\alpha_k = \exp\!\bigl(f(k-1)\bigr)$ with $f = \ln(2)/K$\;
%     \ForEach{$x_i \in \mathcal{P}_{k-1}$}{
%         Generate candidate set $C_{x_i}$ from $\mathcal{M}$ under $\mathcal{T}$\;
%         Compute $\text{DFS}(x_i)$ and $\text{DFS}^{*}(x_i)$\;
%         Compute fitness $\mathcal{F}(x_i)$ using $\alpha_k$\;
%     }
%     Select top $M$ candidates with highest fitness to form parent set $\mathcal{S}_k$\;
%     Generate offspring via crossover and mutation (rate $p_m$) to form new population $\mathcal{P}_k$\;
% }
% Return $x' = \arg\max_{x_i \in \mathcal{P}_K} \mathcal{F}(x_i)$\;
% \caption{Diversity-Guided Genetic Attack}
% \label{alg:diversity_attack}
% \end{algorithm}

\looseness-1\textbf{Remark.}
Our design leverages two key observations: (i) TTS strategies are highly dependent on candidate diversity since they rely on aggregating or scoring multiple generations, and (ii) early-stage guidance (via $\text{DFS}^{*}$) prevents premature convergence and helps the stress test population reach promising regions of the prompt space. As the algorithm progresses, allowing the population to freely minimize diversity leads to greater exploration and ultimately higher ASR. This resembles a \textit{curriculum learning} \cite{bengio2009curriculum} approach where the adversary first \emph{teaches} the model to move toward unsafe completions and then lets the optimization converge flexibly, exhibiting a key failure mode of TTS strategies. The algorithm protocol is visualized in Figure \ref{fig:RefDiv}.

% \ac{Nahin to add figure}

% \begin{figure}[t]
%     \centering
%     \includegraphics[width=1.0\linewidth]{diversity_figures/RefDiv-Final.png}\vspace{-4mm}
%     \caption{Progression of RefDiv across iterations and change in response diversity.}
%     \label{fig:RefDiv}
% \end{figure}

%\subsection{Why Does the Attack Work?}

\section{Experiments and Results}

%\ac{Nahin to add to whole section}

\subsection{Experimental Setup}

\looseness-1\textbf{LLMs and Dataset.}
In our experiments, we primarily employ LLMs across different sizes and types: Mistral-7B \citep{jiang2023mistral7b}, Llama3.1-8B \citep{grattafiori2024llama3herdmodels}, Qwen3-8B \citep{yang2025qwen3technicalreport}, and Gemma3-27B \citep{gemmateam2025gemma3technicalreport}. Among these, Mistral-7B and Llama3.1-8B are pure text-based LLMs, Qwen3-8B is a text-based reasoning LLM, and Gemma3-27B is a multimodal LLM. We have also extended our experiments to Llama3.1-70B \citep{grattafiori2024llama3herdmodels}, Phi-4-mini \citep{microsoft2025phi4minitechnicalreportcompact}, Zephyr-7b-r2d2 \citep{tunstall2023zephyrdirectdistillationlm}, and Vicuna-1.5-7b \citep{zheng2023judgingllmasajudgemtbenchchatbot}. For closed-source LLMs, we employ GPT-4.1, o3-mini, Gemini-2.5-Flash, Gemini-2.5-Pro, and Claude-3.5-Haiku. %Although they have different capabilities, we used them only as text-based LLMs.
To evaluate our stress test alongside adversarial attack strategies, we use the popular AdvBench \citep{zou2023universal} benchmark dataset, designed to evaluate the safety-alignment of LLMs by probing how they respond to adversarial instructions. AdvBench contains 520 adversarial queries and corresponding potential harmful responses across diverse domains including cybersecurity, misinformation, fraudulent activities, hate speech, among others.

\textbf{TTS Strategies.} 
In our experiments, we employ two popular baseline TTS strategies: Best-of-$N$ and Monte Carlo Tree Search (MCTS). Best-of-$N$ generates $N$ candidate responses and scores them via a reward model to select the best candidate. We conduct experiments with three reward models for this purpose: \textit{PairRM} \citep{llm-blender-2023}, \textit{deberta-v3-large-v2} by OpenAssistant \citep{he2023debertav3} and \textit{ToxiGuardRail} \citep{nicholas22aira} (additional details on reward models are provided in Appendix \ref{app:guardrail_details}). In experiments, we also vary $N = 2,8,16$. For MCTS, we utilize the open-source implementation provided in the \textit{llm-mcts-inference}\footnote{\textit{https://pypi.org/project/llm-mcts-inference/}} package. Moreover, each instantiation is run with default parameters for the number of children (=3), for a total of 3 MCTS iterations. We also consider a smaller configuration with two children and two iterations (for additional details on MCTS, please see Appendix \ref{app:algo_details_mcts}).

\looseness-1\textbf{Baselines and Evaluation.}
We compare \textsc{RefDiv} with three state-of-the-art jailbreak attack baselines: Greedy Coordinate Gradient (GCG) \citep{zou2023universaltransferableadversarialattacks}, AutoDAN \citep{liu2024autodangeneratingstealthyjailbreak}, and AutoDAN-Turbo \citep{liu2024autodan}. Following the standard AutoDAN evaluation protocol, we evaluate GCG, AutoDAN, AutoDAN-Turbo, and \textsc{RefDiv} using Attack Success Rate (ASR), by measuring ASR for adversarial stress test strings that lead to harmful LLM generations.

%GCG is a white-box attack that optimizes short token suffixes using gradient signals, producing compact prompts with high transferability across models. GCG does not focus on input query interpretation and generates random sequences that increase the probability of generating particular harmful responses. In contrast, PAIR is a black-box attack where an attacker LLM iteratively proposes and refines jailbreak prompts based on the target model's responses, achieving query-efficient and natural jailbreaks. AutoDAN is an improved gradient-based method like GCG but generates interpretable and human-like adversarial prompts, balancing attack success with readability for better transferability. It uses a GA to gradually improve the input queries based on a gradient-based fitness function. Table \ref{tab:attck_strategies} summarizes the comparison among the baseline strategies.

% \begin{table}[h]
% \centering
% \begin{tabular}{lccp{6cm}}
% \hline
% \textbf{Attack} & \textbf{Access} & \textbf{Output Style} & \textbf{Key Idea} \\ \hline
% GCG & White-box & Random suffixes & Gradient-guided token optimization to craft short adversarial suffixes. \\ 
% PAIR & Black-box & Natural prompts & Iteratively refines jailbreak prompts using feedback from the target model. \\
% AutoDAN & White-box & Natural prompt template & Generates human-readable adversarial prompts by balancing attack success with naturalness. \\ \hline
% \end{tabular}
% \caption{Comparison of GCG, PAIR, and AutoDAN LLM attack strategies.}
% \label{tab:attck_strategies}
% \end{table}

\subsection{Main Results}

\begin{figure*}[t]
    \centering
    \includegraphics[width=0.8\linewidth]{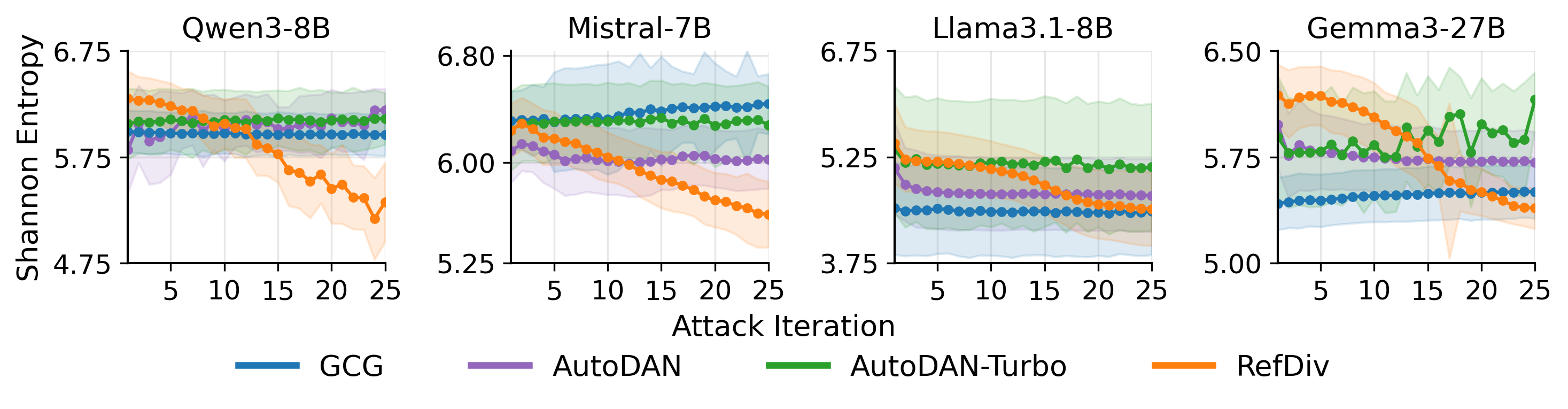}\vspace{-4mm}
    \caption{Analyzing the Shannon Entropy trend across iterations for \textsc{RefDiv} and AutoDAN.}
    \label{fig:shannon-comparison-bon}\vspace{-4mm}
\end{figure*} %\vspace{-2mm}

\looseness-1We compare \textsc{RefDiv} with our three baseline methods to demonstrate how it uncovers the diversity-dependence of TTS, eventually leading to significant output failure. Table~\ref{tab:asr_results_bon} presents the Attack Success Rate (ASR) of the attack methods on TTS with Best-of-$N$ ($N=8$ and reward model: \textit{PairRM}) and MCTS (children: 3, iterations: 3) across multiple models. We showcase the ASR trend over iterations for each attack across LLMs and TTS strategies in Figure \ref{fig:main-asr-comparison-bon} (Best-of-$N$) and Figure \ref{fig:main-asr-comparison-mcts} (MCTS). 

\looseness-1For Best-of-$N$, \textsc{RefDiv} consistently outperforms other methods, achieving more than 7\% ASR margin for Llama3.1-8B and over a 17\% margin for Gemma3-27B. This trend showcases the failure mode and diversity-sensitive nature of TTS strategies. Similarly, for Mistral-7B, \textsc{RefDiv} outperforms other methods, although for Qwen3-8B \textsc{RefDiv} has a lower ASR (0.995) to AutoDAN (0.996) with only a difference of 0.001. Furthermore, \textsc{RefDiv} outperforms AutoDAN for Llama3.1-70B, Phi-4-mini, Zephyr-7b-r2d2, and Vicuna-1.5-7b with significant margins. Additional results for these models are provided in Appendix \ref{app:extended_experiments_model}. 

In Best-of-$N$, AutoDAN-Turbo achieves 0.07 to 0.78 lower ASR than other methods showing more inconsistency in performance. This gap illustrates the limitation of standard API-based attacks that ignore post-generation selection, and highlights the robustness of \textsc{RefDiv}'s diversity-targeting approach in TTS settings. Additionally, even though AutoDAN-Turbo employs a lifelong learning agent pre-trained on harmful query subsets, giving it an inherent advantage through prior exposure to malicious distributions, it is not very performant for TTS. In contrast, \textsc{RefDiv} is entirely training-free and operates solely at inference time which makes \textsc{RefDiv} more practical. GCG shows limited effectiveness in TTS and underperforms significantly for almost all baselines and models.

For MCTS, \textsc{RefDiv}'s stress test results in a major degradation of TTS performance compared to baselines: for Qwen3-8B and Mistral-7B both AutoDAN and \textsc{RefDiv} attain perfect ASR (1.0) but \textsc{RefDiv} achieves significant ASR margins compared to AutoDAN for both Llama3.1-8B and Gemma3-27B. Specifically, for Llama3.1-8B \textsc{RefDiv} attains 0.967 ASR compared to AutoDAN's 0.831 and for Gemma3-27B \textsc{RefDiv} achieves 0.989 compared to AutoDAN's 0.904. Interestingly, we find that \textsc{RefDiv} shows reduced sensitivity to MCTS hyperparameters and attains consistently strong performance (additional results provided in Appendix \ref{app:sensitivity-mcts}, which demonstrate this phenomenon). GCG achieves almost a perfect ASR similar to AutoDAN and \textsc{RefDiv} for Mistral-7B. However, it does not generalize well to other models. AutoDAN-Turbo does not work well for TTS, potentially because the default distribution of the agent's skill library might not align well with the TTS reasoning stage. For example, on Gemma3-27B, AutoDAN-Turbo achieves an ASR of only 0.156, whereas \textsc{RefDiv} achieves the highest ASR of 0.989.

% \begin{table}[t]
%     \centering
%     \caption{Comparing ASR of \textbf{\textsc{RefDiv} (Ours)} and baselines: \textbf{GCG}, AutoDAN (\textbf{AD}), and AutoDAN-Turbo (\textbf{ADT}), with the best performer shown in \textcolor{red!75!black}{red}.}
%     \label{tab:asr_results_bon}
%     \vspace{-2mm}
%     \footnotesize % Slightly smaller than \small to ensure fit
%     \setlength{\tabcolsep}{3pt} % Tighten horizontal spacing
%     \resizebox{0.43\textwidth}{!}{%
%     \begin{tabular}{llccccc}
%         \toprule
%         \textbf{TTS} & \textbf{Model} & \textbf{GCG} & \textbf{AD} & \textbf{ADT}  & \textbf{\textsc{RefDiv}} \\ 
%         \midrule
%         Best-of-$N$ & Qwen3-8B    & 0.335 & \textcolor{red!75!black}{{0.996}} & 0.465  & 0.995 \\
%         ($N=8$)     & Mistral-7B  & 0.877 & 0.973 & 0.733  & \textcolor{red!75!black}{{0.976}} \\ 
%                     & Llama3.1-8B & 0.176 & 0.368 & 0.395  & \textcolor{red!75!black}{{0.465}} \\ 
%                     & Gemma3-27B  & 0.054 & 0.749 & 0.171 & \textcolor{red!75!black}{{0.926}} \\ 
%         \midrule
%         MCTS        & Qwen3-8B    & 0.400 & \textcolor{red!75!black}{{1.000}} & 0.500  & \textcolor{red!75!black}{{1.000}} \\
%                     & Mistral-7B  & 0.996 & \textcolor{red!75!black}{{1.000}} & 0.748 & \textcolor{red!75!black}{{1.000}} \\ 
%                     & Llama3.1-8B & 0.254 & 0.831 & 0.446 & \textcolor{red!75!black}{{0.967}} \\ 
%                     & Gemma3-27B  & 0.336 & 0.904 & 0.148 & \textcolor{red!75!black}{{0.989}} \\ 
%         \bottomrule
%     \end{tabular}}
% \vspace{-7mm}
% \end{table}

\begin{table}[t]
    \centering
    \caption{Comparing ASR of \textbf{\textsc{RefDiv} (Ours)} and baselines: \textbf{GCG}, AutoDAN (\textbf{AD}), and AutoDAN-Turbo (\textbf{ADT}), with the best performer highlighted in \hl{red}.}
    \label{tab:asr_results_bon}
    \vspace{-2mm}
    \footnotesize % Slightly smaller than \small to ensure fit
    \setlength{\tabcolsep}{3pt} % Tighten horizontal spacing
    \resizebox{0.43\textwidth}{!}{%
    \begin{tabular}{llccccc}
        \toprule
        \textbf{TTS} & \textbf{Model} & \textbf{GCG} & \textbf{AD} & \textbf{ADT}  & \textbf{\textsc{RefDiv}} \\ 
        \midrule
        Best-of-$N$ & Qwen3-8B    & 0.335 & \cellcolor{red!10}{{0.996}} & 0.414  & 0.995 \\
        ($N=8$)     & Mistral-7B  & 0.877 & 0.973 & 0.733  & \cellcolor{red!10}{{0.976}} \\ 
                    & Llama3.1-8B & 0.176 & 0.368 & 0.397  & \cellcolor{red!10}{{0.465}} \\ 
                    & Gemma3-27B  & 0.054 & 0.749 & 0.171 & \cellcolor{red!10}{{0.926}} \\ 
        \midrule
        MCTS        & Qwen3-8B    & 0.400 & \cellcolor{red!10}{{1.000}} & 0.485  & \cellcolor{red!10}{{1.000}} \\
                    & Mistral-7B  & 0.996 & \cellcolor{red!10}{{1.000}} & 0.764 & \cellcolor{red!10}{{1.000}} \\ 
                    & Llama3.1-8B & 0.254 & 0.831 & 0.446 & \cellcolor{red!10}{{0.967}} \\ 
                    & Gemma3-27B  & 0.336 & 0.904 & 0.156 & \cellcolor{red!10}{{0.989}} \\ 
        \bottomrule
    \end{tabular}}
\vspace{-7mm}
\end{table}

\looseness-1Note that the limited success of GCG can be attributed to its use of a comparatively weaker optimizer and a singular focus on the final output of the LLM, neglecting the internal effects of diverse candidate selection guided by a reward model or via MCTS. In comparison to AutoDAN or AutoDAN-Turbo, which do not seek to constrain TTS candidate diversity, \textsc{RefDiv} minimizes token-level diversity via Shannon Entropy while constraining the model to harmful generations, thus effectively exposing the failure mode of TTS strategies. 

%\textcolor{red}{We have also extended our evaluation to state-of-the-art jailbreak strategies, including AutoDAN-Turbo and MouseTrap. As detailed in Appendix~\ref{app:sota_baselines}, \textsc{RefDiv} consistently outperforms both methods. To additionally assess robustness, we evaluate \textsc{RefDiv} on a broader set of models: Llama3.1-70B, Phi-4-mini, Zephyr-7b-r2d2, and Vicuna-1.5-7b. Appendix~\ref{app:extended_experiments_model} demonstrates that \textsc{RefDiv} maintains strong performance across this diverse model suite.}

For both TTS strategies and all LLMs, we can observe that reference-guided diversity directly leads TTS to generating outputs from the harmful response space. In particular, for LLMs such as Llama3.1-8B and Gemma3-27B where other methods fail, the \textsc{RefDiv} stress test works well. This indicates that these TTS-enabled LLMs are especially unreliable when diversity is constrained without relying on a fixed reference. We provide additional experiments for $N=2,16$ in Appendix \ref{sec:variable-n}. 
%\textcolor{red}{Additionally, we vary the hyperparameters of the MCTS component to assess the stability of our search procedure. As shown in Appendix~\ref{app:sensitivity-mcts}, \textsc{RefDiv} remains consistently effective across alternative MCTS configurations, indicating that its performance does not rely on specific tuning choices.}. 

%We further analyzed ASR progression across iterations for AutoDAN and \textsc{RefDiv} only, as GCG exhibited consistently low ASR. Figure~\ref{fig:main-asr-comparison-bon} shows ASR growth over iterations for best-of-n(8), and Figure~\ref{fig:main-asr-comparison-mcts} shows ASR progression for MCTS. 

%To investigate the robustness of RefDiv we have further conducted multiple experiments by varying the number of candidates in best-of-n and reward model. Appendinx \ref{sec:variable-n}  and \ref{sec:variable-reward} discusses briefly about those experiments and reults. 

% \begin{figure}[h]
%     \centering
%     \includegraphics[width=1.0\linewidth]{diversity_figures/shannon_entropy_comparison_mcts.png}
%     \caption{Comparison of Shannon Entropy across the iterations for MCTS. }
%     \label{fig:shannon-comparison-mcts}
% \end{figure}

\subsection{Why Does \textsc{RefDiv} Work?}

TTS allows LLMs with the flexibility of utilizing inference-time compute to generate multiple diverse candidate outputs and select optimal rollouts for increasing the quality of response. Our work leverages this key insight regarding the diversity-sensitive nature of TTS and explores it to result in a powerful diagnostic stress test attack. Furthermore, in comparison, non-diversity-optimizing attack algorithms such as GCG, AutoDAN, and AutoDAN-Turbo, generally exhibit lower performance compared to our proposed \textsc{RefDiv}. Thus, to analyze why \textsc{RefDiv} works, we plot the candidate token-level Shannon entropy $H$ in the Best-of-$N$ ($N=8$) setting over each iteration in Figure \ref{fig:shannon-comparison-bon}. The figure demonstrates that for RefDiv, Shannon entropy decreases as iterations increase. Interestingly, in the initial iterations, the Shannon entropy for \textsc{RefDiv} is higher than the Shannon entropy for GCG, AutoDAN and AutoDAN-Turbo. As iterations increase, an inversion occurs and the Shannon entropy decreases significantly for \textsc{RefDiv} whereas it remains constant for other methods throughout. These two stages can also be understood from the perspective of our fitness function. In initial iterations for low $t$, owing to the dynamic weighting via $\alpha_t$, the fitness function is primarily driven by the reference-guided diversity score. This guides the GA to follow a particular reference path where the goal is to maximize the likelihood to generate affirmative/reference response tokens. However, in later iterations as $t$ increases (and $\alpha_t$ exponentially increases), \textsc{RefDiv} switches to fully minimizing diversity, thus steering the LLM to converge on some set of harmful responses. This hybrid approach of exploitation-exploration makes \textsc{RefDiv} significantly more robust than other stress test methods and reveals the inherent diversity-sensitive failure mode of TTS.

\textbf{Remark.} Owing to space constraints, we provide the diversity trends for MCTS in Appendix \ref{sec:shanon-mcts}. Moreover, we also evaluate alternative increasing weighting schedules for $\alpha(t)$ (results in Appendix~\ref{app:sensitivity-alpha}) and observe consistently similar performance across all variants, implying low sensitivity to parametric choices. Finally, our additional quantitative analysis in Appendix~\ref{app:entropy_corr_analysis} reveals that TTS pipelines are highly sensitive to \textit{diversity suppression} and that ASR exhibits \textit{strong negative correlation} with entropy, reinforcing the central role of diversity in maintaining TTS safety.

\begin{figure}[t] 
    \centering
    \includegraphics[width=0.4\textwidth]{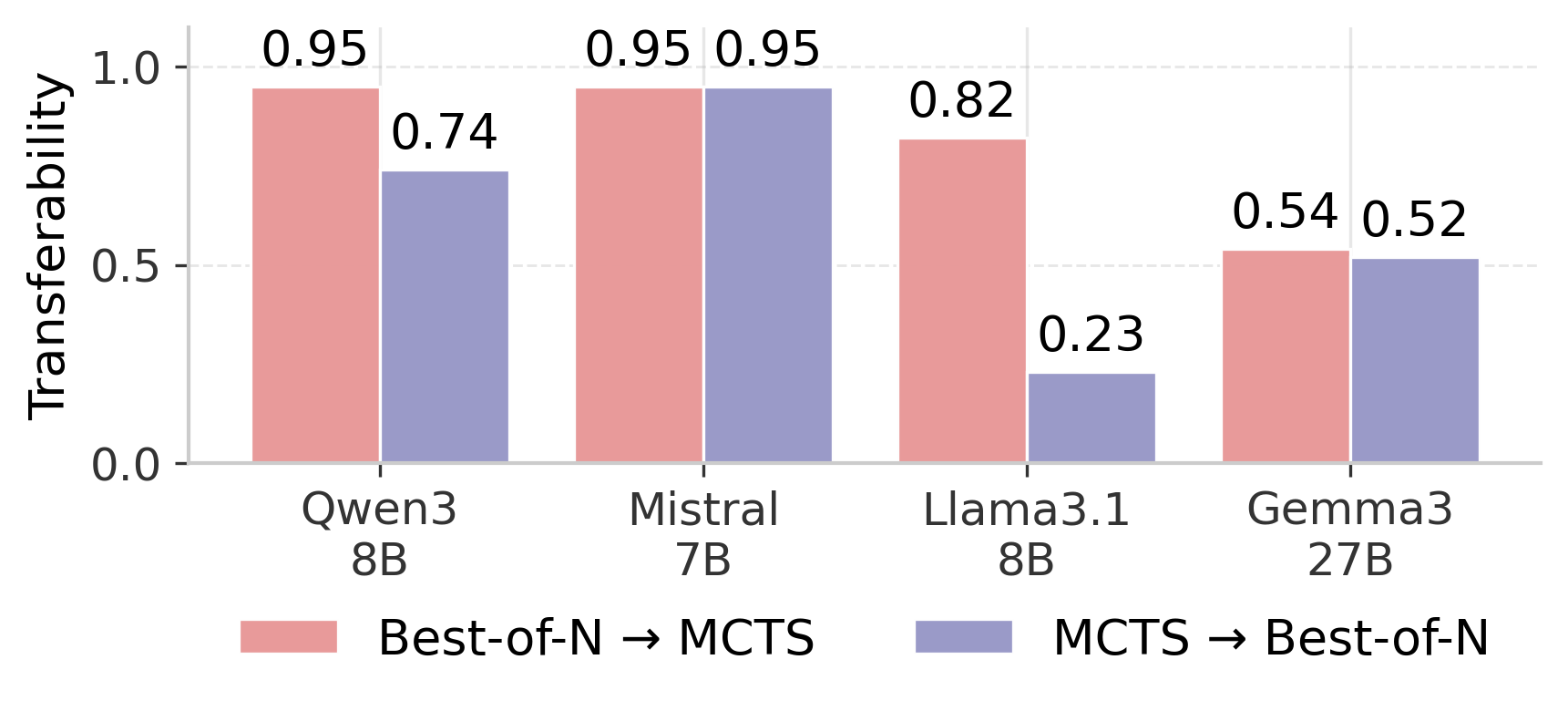}\vspace{-3mm}
    \caption{Transferability of \textsc{RefDiv} prompts for Best-of-$N$ $\rightarrow$ MCTS and MCTS $\rightarrow$ Best-of-$N$ across LLMs. 
    %\sn{Don't like the color combo here, please make it different}
    }\label{fig:transfer-tts}\vspace{-5mm}
\end{figure}

\subsection{Black-Box Transferability Across TTS Strategies}
\label{subsec: trans-tts}

\looseness-1Clearly, while \textsc{RefDiv} works well for the white-box setting, a natural subsequent question to answer is: \textit{do adversarial prompts generated for a specific TTS strategy by \textsc{RefDiv} transfer across different TTS strategies?} Essentially, in this case the adversary is aware of the target LLM being used, but not the specific TTS strategy employed by them. Moreover, if adversarial strings can transfer across TTS strategies, this explicitly indicates that the diversity-specific failure mode of TTS is a fundamental property of TTS frameworks, and not arising only due to the LLM. 
To analyze this, we quantify the ASR for how \textsc{RefDiv} Best-of-$N$ (MCTS) prompt samples transfer to MCTS (Best-of-$N$) across each LLM, and vice versa. These results are provided in Figure \ref{fig:transfer-tts}. Interestingly, for Mistral-7B and Gemma3-27B the results demonstrate that our adversarial stress test strings crafted for one TTS strategy remain similarly effective for the other. However, for Qwen3-8B and Llama3.1-8B, transferability from Best-of-$N$ $\rightarrow$ MCTS is notably higher than the transferability from MCTS $\rightarrow$ Best-of-$N$.
%We analyzed the performance of adversarial attacks on the best-of-n(8) strategy using adversarial strings generated by attacking MCTS, and vice versa. Figure \ref{fig:transfer-tts} illustrates the transferability of RefDiv across different TTS strategies and for fixed models. The results show that adversarial strings crafted against one TTS strategy remain similarly effective when transferred to another strategy in the cases of Mistral-7B and Gemma3-27B. In contrast, for Qwen3-8B and Llama3.1-8B, transferability from best-of-n(8) to MCTS is notably higher than the reverse. Overall, these findings demonstrate that RefDiv is transferable across TTS strategies, indicating its capability to generate strong adversarial strings that consistently compromise LLM security.

\subsection{Black-Box Transferability To Closed-Source LLMs}
\label{sec:trans-open-source}

\begin{figure}[t] %{r}{0.65\textwidth}
    \centering
    \resizebox{0.48\textwidth}{!}{
    \begin{subfigure}{0.5\textwidth}
        \centering
        \includegraphics[width=\linewidth]{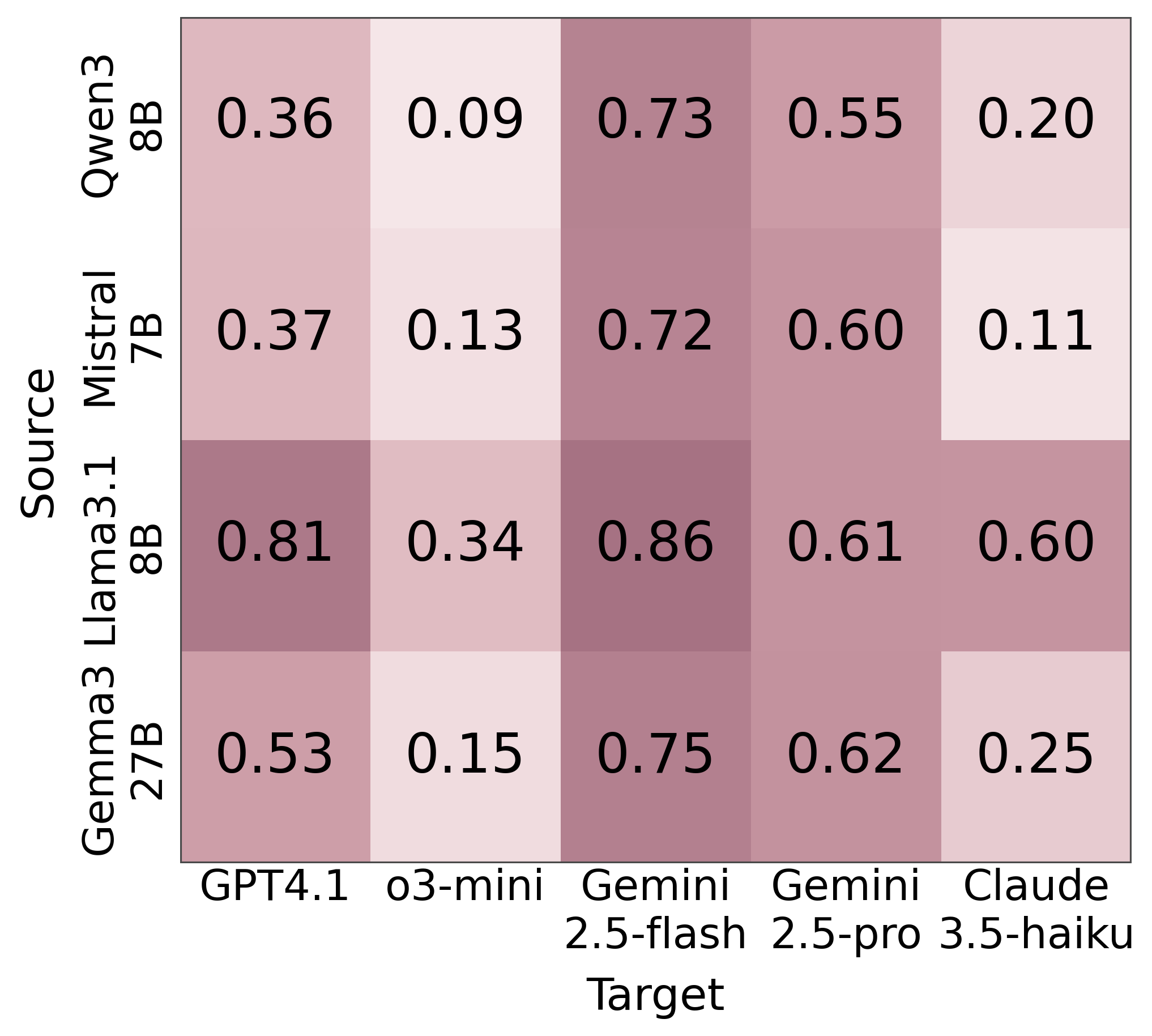}\vspace{-3mm}
        %\caption{Best-of-$N$}
        \label{fig:transfer-bon-gpt}
    \end{subfigure}%
    \hfill
    \begin{subfigure}{0.5\textwidth}
        \centering
        \includegraphics[width=\linewidth]{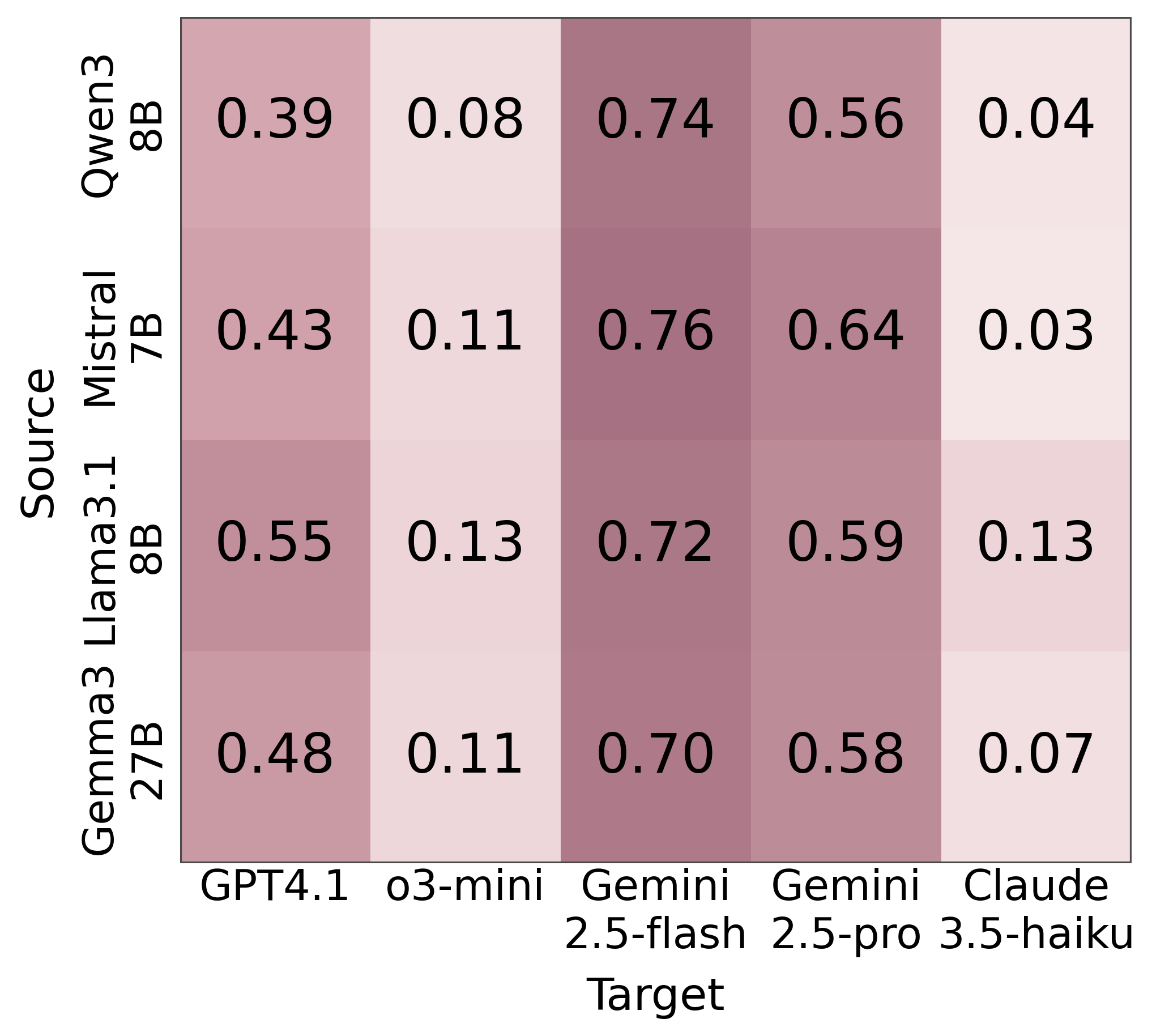}\vspace{-3mm}
        %\caption{MCTS}
        \label{fig:transfer-bon-gpt-mcts}
    \end{subfigure}}\vspace{-3mm}
    \caption{Black-box transferability (ASR) of \textsc{RefDiv} from open-source to closed-source LLMs. Best-of-$N$ (\textit{left}) and MCTS (\textit{right}).}
    \label{fig:transfer-gpt}
\vspace{-6mm}
\end{figure}

\looseness-1Even more importantly, while \textsc{RefDiv} generated prompts transfer well across TTS strategies, the previous scenario assumes the LLM models are known to the adversary. Thus, we now relax this assumption and assume only \textit{black-box} input-output access to the LLM, leading us to ask: \textit{do the adversarial stress test prompts generated by \textsc{RefDiv} transfer across closed-source LLMs as well?} We thus investigate the transferability of \textit{successful} prompts generated using \textit{source} (open-source) LLMs to \textit{target} closed-source models: GPT-4.1, o3-mini (reasoning), Gemini-2.5-Flash (reasoning), Gemini-2.5-Pro (reasoning), and Claude-3.5-Haiku. 

Our findings in Figure \ref{fig:transfer-gpt} demonstrate that successful queries generated on Llama3.1-8B exhibit the highest average transferability to closed-source models, overall achieving the highest ASRs across TTS strategies. We also undertake a qualitative analysis of \textsc{RefDiv} attack queries for Llama3.1 in Appendix \ref{app:qualitative_transfer} to uncover linguistic patterns that contribute to the success of this attack. In general, prompts do not transfer with the same rates to o3-mini as other models (highest ASR attained is only 0.34 using Llama3.1-8B and Best-of-$N$), although this is still a significant success rate. Moreover, Gemini-2.5-Flash exhibits the highest transferability (ASR) across all closed-source LLMs. Our results thus show that \textsc{RefDiv} attacks can be employed in a fully black-box setting where closed-source LLMs are the targets. %We undertake a qualitative analysis of \textsc{RefDiv} generated queries in Appendix \ref{app:qualitative_transfer} 
% \sn.

% 0.4025, 0.485, 0.525, 0.478

% \begin{wrapfigure}{r}{0.4\textwidth}
%     \centering
%     \includegraphics[width=0.45\textwidth]{diversity_figures/transfer-bon-gpt.png}
%     \caption{Transferability of RefDiv (best-of-n) to OpenAI models.\ac{Increase font size of model names (and Source/Target) a lot more and decrease font size of cell text quite a bit more so it is similarly sized to main paper text font size. Also put a border around the heatmap. This goes for all the heatmaps}}
%     \label{fig:transfer-bon-gpt}
% \end{wrapfigure}

% \begin{wrapfigure}{r}{0.4\textwidth}
%     \centering
%     \includegraphics[width=0.45\textwidth]{diversity_figures/transfer-bon-gpt-mcts.png}
%     \caption{Transferability of RefDiv (MCTS) to OpenAI models.}
%     \label{fig:transfer-bon-gpt-mcts}
% \end{wrapfigure}

\looseness-1\textbf{Remark.} As Table \ref{tab:asr_results_bon} shows, \textsc{RefDiv} achieves significantly higher ASR for Qwen-3-8B and Mistral-7B compared to other models. These models can therefore be considered more \textit{susceptible} to adversarial prompts, and end up generating weaker queries that demonstrate limited transferability to potentially more robust closed-source LLMs. In contrast, Llama3.1-8B and Gemma3-27B exhibit greater resistance to adversarial inputs, necessitating the generation of more sophisticated queries for harmful response generation, and in turn exhibiting significantly higher transferability to closed-source LLMs. %\textcolor{red}{We have provided a qualitative analysis of transferability in Appendix \ref{app:qualitative_transfer}}. 
Overall, \textsc{RefDiv} generates prompts that transfer successfully across all the closed-source reasoning and non-reasoning LLMs.%, underscoring the impact of our proposed method. 

% \textbf{Black-Box TTS Setting.} In practical deployments, adversaries typically lack direct access to the target's specific TTS pipeline ($\mathcal{M}$,$\mathcal{T}$). However, our results indicate that the reliance on candidate diversity is a fundamental vulnerability of TTS strategies, enabling effective attacks via surrogate models like Llama3.1-8B. By optimizing against the candidate distribution of a local, open-source TTS system, adversaries can craft prompts that induce similar failures in black-box environments. Transferability between TTS systems shown in \cref{subsec: trans-tts} also confirms the effectiveness of surrogate models in Black-Box settings.

\subsection{Potential Mitigation Strategies}
%Given the observed failure modes of TTS systems under adversarial prompting, in this section we investigate several defenses. We focus on mitigation strategies that are commonly used in practi

Given \textsc{RefDiv}'s success against TTS-enabled LLMs, we now study several potential mitigation strategies, including standard approaches such as (a) perplexity-based filtering \cite{jain2023baseline}, (b) utilizing safety-specific reward models, (c) increasing the candidate diversity for TTS to help counter \textsc{RefDiv}'s diversity reduction objective, and (d) employing state-of-the-art safety guardrail classifiers.

\subsubsection{Perplexity Filtering}
\label{app:mitigation-perplexity}

\looseness-1Prior work has utilized perplexity-based filtering to ascertain whether adversarial prompts consist of strings that are incoherent and generated using an optimization procedure \cite{jain2023baseline}. While this defense is quite primitive, and only works well against simple attacks such as GCG, we conduct an experiment to assess whether it is an effective potential mitigation strategy for \textsc{RefDiv}. We consider Llama3.1-8B as the target model, given its exceptional transferability (as evidenced in Figure \ref{fig:transfer-gpt}). Then, for each attack strategy, we pool all the prompts and remove the top-10\% and top-20\% of prompts with highest perplexity (computed using a standalone LLM, Qwen2.5-7B, for fairness). We then count how many total prompts were not filtered for each attack individually, and how many of these are actually successful jailbreaks. Due to space limitations, we provide these results in Appendix \ref{app: llama-perplexity}. As can be observed, for both settings, \textsc{RefDiv} achieves the highest success rate of 42.7\%, while AutoDAN and AutoDAN-Turbo achieve 40.4\% and 39.7\% respectively. This indicates that a majority of attack samples are very low perplexity, thereby invalidating the perplexity defense.

% \begin{table}[t]\small
%     \centering
%     \caption{Average perplexity (PPL) of adversarial prompts. We compare GCG, AutoDAN (\textbf{AD}), AutoDAN-Turbo (\textbf{ADT}), and \textbf{\textsc{RefDiv (Ours)}}. Lower PPL indicates higher text fluency (highlighted in \hl{red}).}
%     \label{tab:perplexity}
%     \small % Keep font size consistent with ICML style
%     \setlength{\tabcolsep}{4pt} % Reduce horizontal padding between columns
%     \resizebox{0.42\textwidth}{!}{
%     \begin{tabular}{lrrrrr}
%         \toprule
%         \textbf{Model} & \textbf{GCG} & \textbf{AD} & \textbf{ADT} & \textbf{\textsc{RefDiv}} \\
%         \midrule
%         Qwen3-8B    & 2800.68 & 119.62 & \cellcolor{red!10}59.83  & 146.28  \\
%         Mistral-7B  & 742.17 &  128.03 &  \cellcolor{red!10}89.45 &  115.25 \\
%         Llama3.1-8B & 1705.51 &  191.11 & \cellcolor{red!10}54.11  &  170.45  \\
%         Gemma3-27B  & 1552.04 &  157.16 &  \cellcolor{red!10}120.66 &  156.80  \\
%         \bottomrule
%     \end{tabular}}
% \vspace{-3mm}
% \end{table}

% To test whether adversarial prompts are easily flagged by perplexity filters, we have measured average perplexity (with Qwen2.5-7B \citep{qwen2025qwen25technicalreport})  for the successful queries. Table~\ref{tab:perplexity} shows that \textsc{RefDiv} maintains low perplexity similar to other methods, whereas gradient-based GCG produces extremely high-perplexity  nonsensical prompts that would be trivially filtered.

\subsubsection{Safety-Specific Reward Models}
\label{sec:variable-reward}

\looseness-1To ensure our results are not reward-specific, we evaluate \textsc{RefDiv} for Best-of-$N$ ($N=8$) using two other safety-aligned reward models: \textit{deberta-v3-large-v2} and \textit{ToxiGuardRail}. We provide results in Appendix \ref{app: experiments with reward models}, demonstrating that \textsc{RefDiv} consistently attains high ASR and outperforms AutoDAN. For instance, on Llama3.1-8B with the \textit{deberta-v3-large-v2} reward model, \textsc{RefDiv} attains a 0.27 ASR compared to AutoDAN's 0.17 ASR. Additionally, ASR and Shannon entropy trends (Figures ~\ref{fig:main-asr-comparison-deberta} and \ref{fig:shannon-comparison-deberta}) closely match those under \textit{PairRM}, showing that stronger safety rewards reduce but do not eliminate diversity-based jailbreaks.

\subsubsection{Increasing Candidate Diversity}
\looseness-1 We seek to analyze whether increasing candidate diversity can potentially counter the diversity-reducing objective of \textsc{RefDiv}. Thus, we conduct experiments where we increase $N$, (the number of candidate responses) for Best-of-$N$ TTS and observe ASR trends. We provide these results in Appendix \ref{sec:variable-n}. As can be observed, simply increasing candidates does not effectively reduce attack performance, instead adding a higher computational overhead.

%the sensitivity of the best-of-$N$ TTS strategy by varying the number of sampled candidates $N$, using \textit{PairRM} as the reward model, as a defense against the diversity reduction mechanism of \textsc{RefDiv}. Appendix \ref{app: experiments with reward models} shows that the performance of \textsc{RefDiv} remains consistent even for higher values of $N$.  
% \sn

\subsubsection{Guardrails/Safety Classifiers}

\begin{figure}[t]
    \centering
    \resizebox{0.48\textwidth}{!}{
    \begin{subfigure}{0.5\textwidth}
        \centering
        \includegraphics[width=\linewidth]{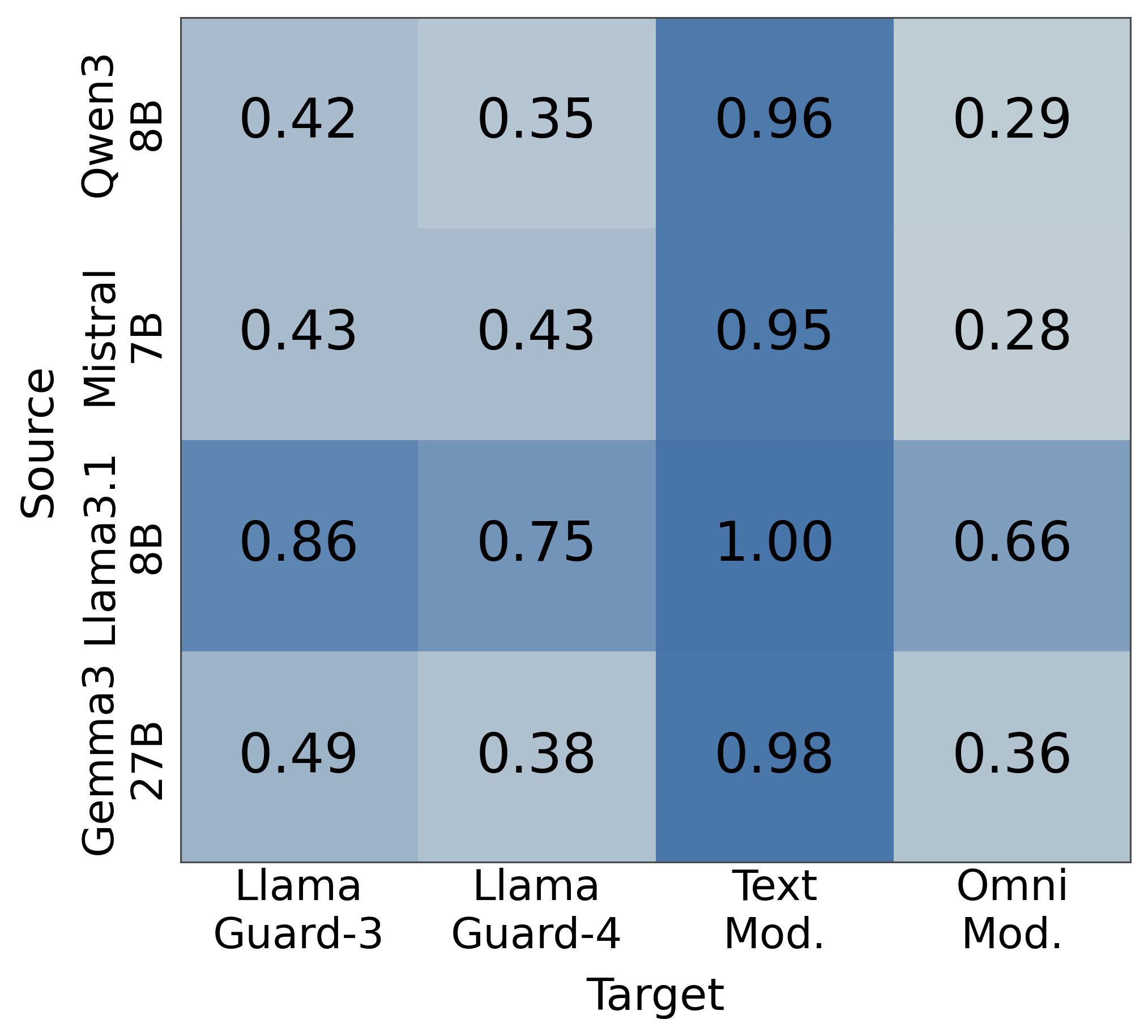}
        %\caption{Best-of-$N$}
        \label{fig:transfer-bon-guard}
    \end{subfigure}%
    \hfill
    \begin{subfigure}{0.5\textwidth}
        \centering
        \includegraphics[width=\linewidth]{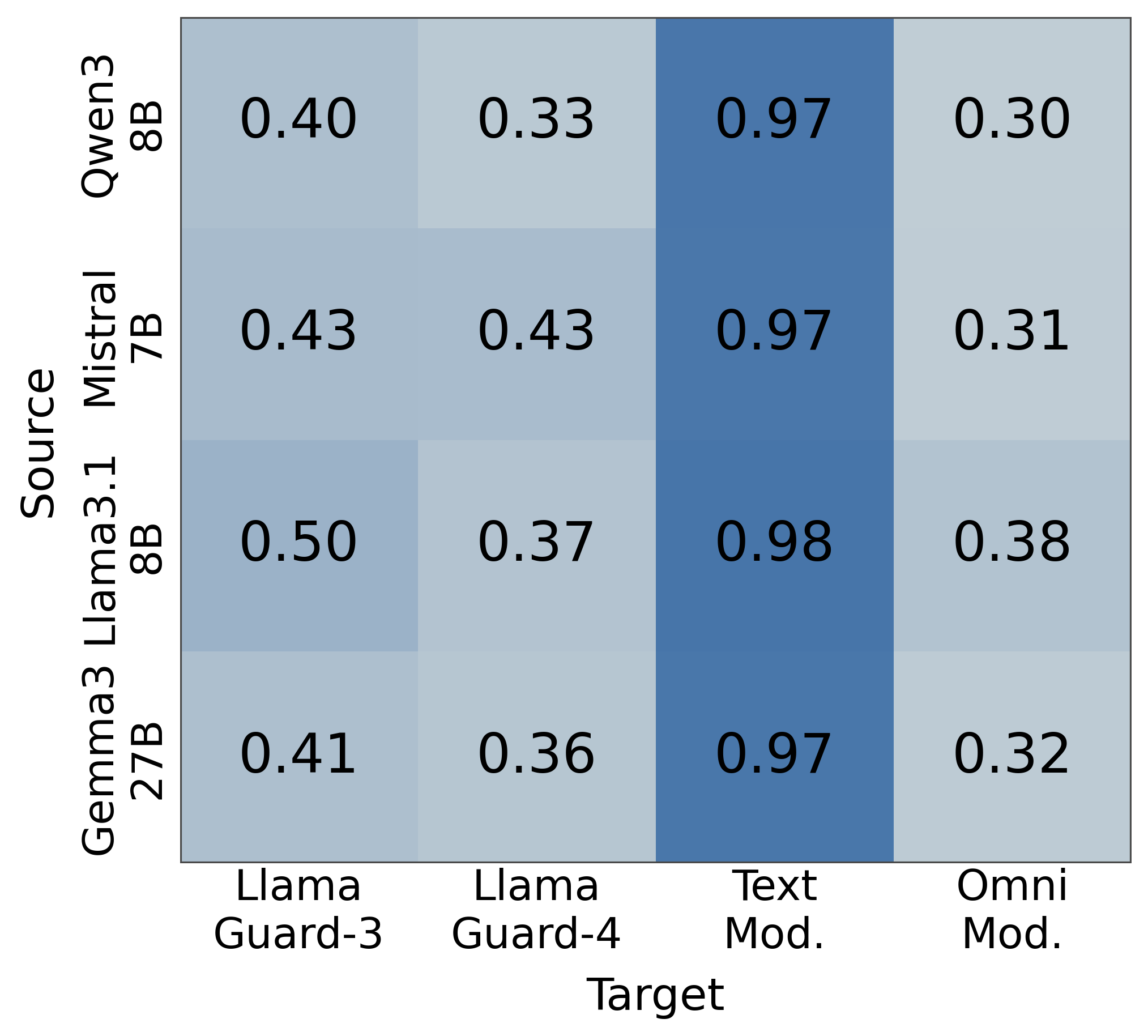}
        %\caption{{MCTS}}
        \label{fig:transfer-bon-guard-mcts}
    \end{subfigure}}\vspace{-3mm}
    \caption{ASR of open-source models attack prompts generated via \textsc{RefDiv} with Best-of-$N$ (\textit{left}) and MCTS (\textit{right}) TTS across several popular guardrail defense classifiers. }
    \label{fig:transfer-guard}\vspace{-8mm}
\end{figure}

\looseness-1Guardrail models are commonly deployed as a first line of defense against adversarial inputs by processing the provided input and filtering/flagging it in case it contains harmful prompt queries. We now seek to analyze if the adversarial prompts generated by \textsc{RefDiv} bypass state-of-the-art guardrail classifiers. If this is the case, guardrails pose limited defensive capability against this diversity-targeted robustness issue exhibited by TTS-based LLMs. We undertake experiments with 4 popular guardrail classifiers: LlamaGuard-3 and LlamaGuard-4 \cite{inan2023llama}, OpenAI Text-Moderation and Omni-Moderation APIs \cite{openai_mod}. We evaluate the robustness of these guardrails against adversarial queries generated by \textsc{RefDiv} for both Best-of-$N$ and MCTS. As illustrated in Figure \ref{fig:transfer-guard}, \textsc{RefDiv}-generated queries are effective at bypassing guardrails, leading to increased false negatives. For instance, for Best-of-$N$, queries generated using Llama3.1-8B successfully transferred to guard models with average ASR $\approx$82\%. The ASR trends for MCTS are similar. Moreover, the strongest adversarial queries are generated using Llama3.1-8B as the source (similar to previous trends), and the OpenAI Text Moderation API exhibits the largest bypass rate compared to the other guardrails. Our findings are also in-line with past work that has found fragility/robustness issues with guardrail classifiers \citep{achara2025watching}.

%In contrast, the ASR under MCTS was around 56\%. The remaining guard models exhibited lower transferability compared to Llama3.1-8B, consistent with previously observed transferability trends for GPT-based models.

% 0.4025, 0.485, 0.525, 0.478

% \begin{wrapfigure}{r}{0.4\textwidth}
%     \centering
%     \includegraphics[width=0.45\textwidth]{diversity_figures/transfer-bon-guard.png}
%     \caption{Transferability of RefDiv (best-of-n) to guard models.}
%     \label{fig:transfer-bon-guard}
% \end{wrapfigure}

% \begin{wrapfigure}{r}{0.4\textwidth}
%     \centering
%     \includegraphics[width=0.45\textwidth]{diversity_figures/transfer-mcts-guard.png}
%     \caption{Transferability of RefDiv (MCTS) to guard models.}
%     \label{fig:transfer-bon-guard-mcts}
% \end{wrapfigure}

%\section{Discussion}

\section{Conclusion}

\looseness-1In this paper, we identified and characterized a novel failure mode unique to Test-Time Scaling (TTS) methods in LLMs, revealing a critical lack of robustness in their \textit{indirect} reliance on candidate diversity. We introduced \textsc{RefDiv}, a reference-guided diversity stress test protocol that induces mode collapse in the candidate response distribution, thereby undermining the robustness benefits typically afforded by TTS. Our extensive experiments demonstrated that \textsc{RefDiv} is effective across multiple TTS strategies, open-source and closed-source models, as well as safety defenses, highlighting the \textit{pervasiveness} and \textit{transferability} of this diversity-specific issue in TTS. These findings underscore the need for future research on diversity-aware TTS systems that maintain the benefits of TTS while mitigating the risk of critical failure due to an overt reliance on candidate diversity. %By exposing this previously overlooked failure mode, our work provides a foundation for developing more robust TTS-based LLM frameworks.

%\clearpage
\section*{Impact Statement}
Our work undertakes stress testing and uncovers a novel candidate-diversity-specific failure mode of TTS-enabled LLMs with the sole aim of improving their safety and robustness. These findings motivate the development of robust, diversity-aware TTS strategies to mitigate the widespread risks associated with TTS.

% \section{Reproducibility Statement}
% % We make our work fully reproducible by releasing the code. The benchmarking data used to evaluate the models is publicly available. All code is released under an open-source license and is hosted in an anonymous repository.
% We provide our code and implementation in an open-source repository: \url{https://anonymous.4open.science/r/RefDiv-57DB/}. All the experiments were run multiple times, and additional parameters required for reproducibility (e.g. temperature, etc.) are provided both in Appendix \ref{app: impl} and the code repository README. The experiments were conducted on a Linux server with 12x NVIDIA DGX B200 GPUs with 192 GB VRAM/GPU.  

% \section{Ethics Statement}

% Our work undertakes stress testing and uncovers a novel candidate-diversity-specific failure mode of TTS-enabled LLMs with the sole aim of improving their safety and robustness. All experiments were conducted in controlled research environments, and no harmful content generated during stress tests will be shared publicly. We disclose our findings responsibly to the community to raise awareness of this novel failure mode of TTS based on candidate diversity and to encourage the development of robust TTS strategies, similar to past work in the ML/AI robustness literature.

\bibliography{refs}
\bibliographystyle{icml2026}

\clearpage
\appendix
\onecolumn
\section*{Appendix}

%\ac{Nahin to add}

\section{Experiments with Best-of-$N$ for Different Values of $N$}
\label{sec:variable-n}
We conducted experiments by varying the value of $N$ in the best-of-$N$ TTS strategy with \textit{PairRM} reward model. Table~\ref{tab:asr_results_bon-all} reports the ASR of \textsc{RefDiv} and AutoDAN under Best-of-$N$ for $N=2, 8, 16$. The results demonstrate that \textsc{RefDiv} consistently outperforms AutoDAN in most cases. For example, in all of the setups with Llama3.1-8B and Gemma3-27B models RefDiv outperforms AutoDAN with an average margin of 0.13. In other models it shows almost similar or better performance. Furthermore, \textsc{RefDiv} achieves comparable performance across all values of $N$.  

\begin{figure}[H]
    \centering
    \includegraphics[width=1.0\linewidth]{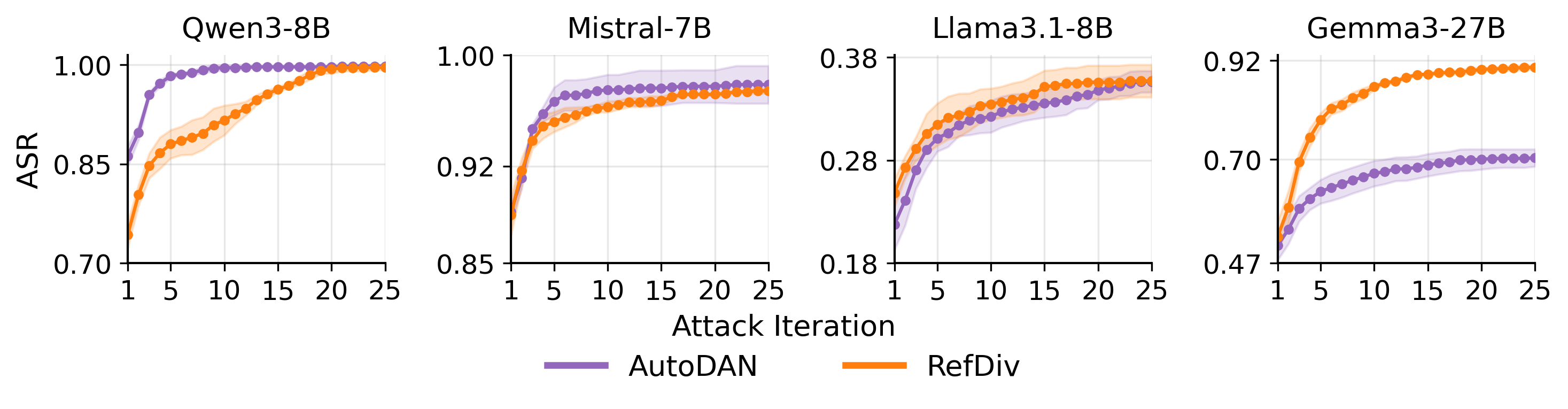}
    \caption{ASR comparison between AutoDAN and \textsc{RefDiv} in Best-of-$N$ TTS ($N=2$).}
    \label{fig:main-asr-comparison-2}
\end{figure}

\begin{figure}[H]
    \centering
    \includegraphics[width=1.0\linewidth]{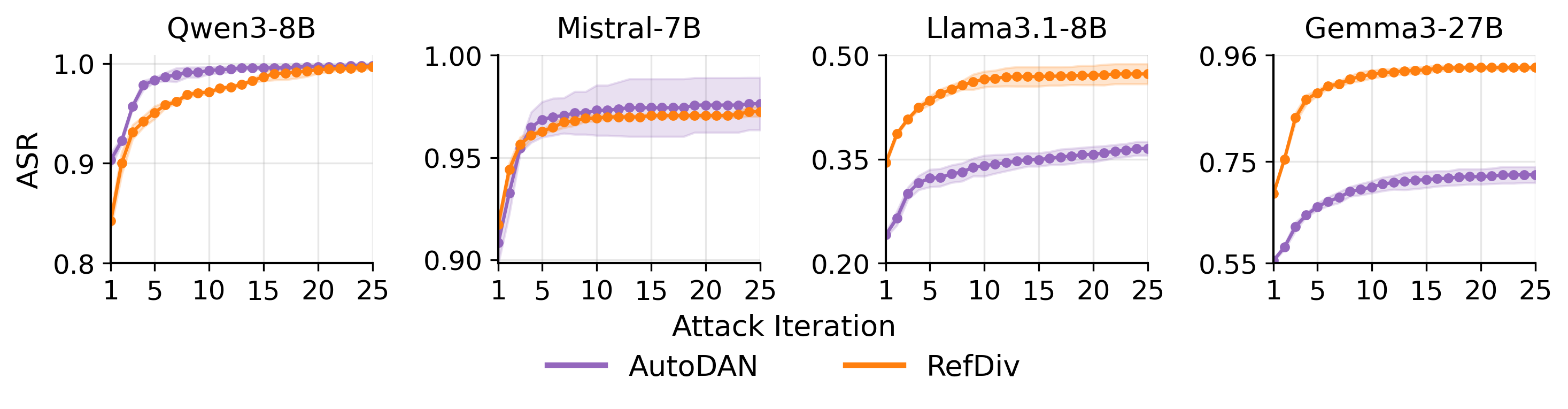}
    \caption{ASR comparison between AutoDAN and \textsc{RefDiv} in Best-of-$N$ TTS ($N=16$).}
    \label{fig:main-asr-comparison-16}
\end{figure}

Figures~\ref{fig:main-asr-comparison-2} and~\ref{fig:main-asr-comparison-16} illustrate the ASR trends for $N$=2 and $N=16$, respectively. For both settings, the ASR curves follow a similar trend to that of $N=8$ for both \textsc{RefDiv} and AutoDAN.  

\begin{table}[H]
\centering
\caption{ASR of different models for various values of $N$ in Best-of-$N$ TTS. The best-performing method is highlighted in \hl{red}.}
\resizebox{0.45\textwidth}{!}{%
\begin{tabular}{llccc}
\hline
\textbf{$N$} & \textbf{Model}  & \textbf{AutoDAN} & \textbf{\textsc{RefDiv} (Ours)} \\ \hline
2   & Qwen3-8B     &  \cellcolor{red!10}{0.998} & 0.996 \\
    & Mistral-7B   & \cellcolor{red!10}{0.979} & 0.974 \\ 
    & Llama3.1-8B  & 0.356 & \cellcolor{red!10}{0.357} \\ 
    & Gemma3-27B   & 0.703 & \cellcolor{red!10}{0.905} \\ \hline
8   & Qwen3-8B     &  \cellcolor{red!10}{0.996} & 0.995 \\
    & Mistral-7B   & 0.973 & \cellcolor{red!10}{0.976} \\ 
    & Llama3.1-8B  & 0.368 & \cellcolor{red!10}{0.465} \\ 
    & Gemma3-27B   & 0.749 & \cellcolor{red!10}{0.926} \\ \hline
16  & Qwen3-8B     &  \cellcolor{red!10}{0.997} & \cellcolor{red!10}{0.997} \\
    & Mistral-7B   &  \cellcolor{red!10}{0.976} & 0.972 \\ 
    & Llama3.1-8B  & 0.365 & \cellcolor{red!10}{0.473} \\ 
    & Gemma3-27B   & 0.724 & \cellcolor{red!10}{0.936} \\ \hline
\end{tabular}
\label{tab:asr_results_bon-all}}
\end{table}

\vspace{-3mm}

\begin{figure}[H]
    \centering
    \includegraphics[width=1.0\linewidth]{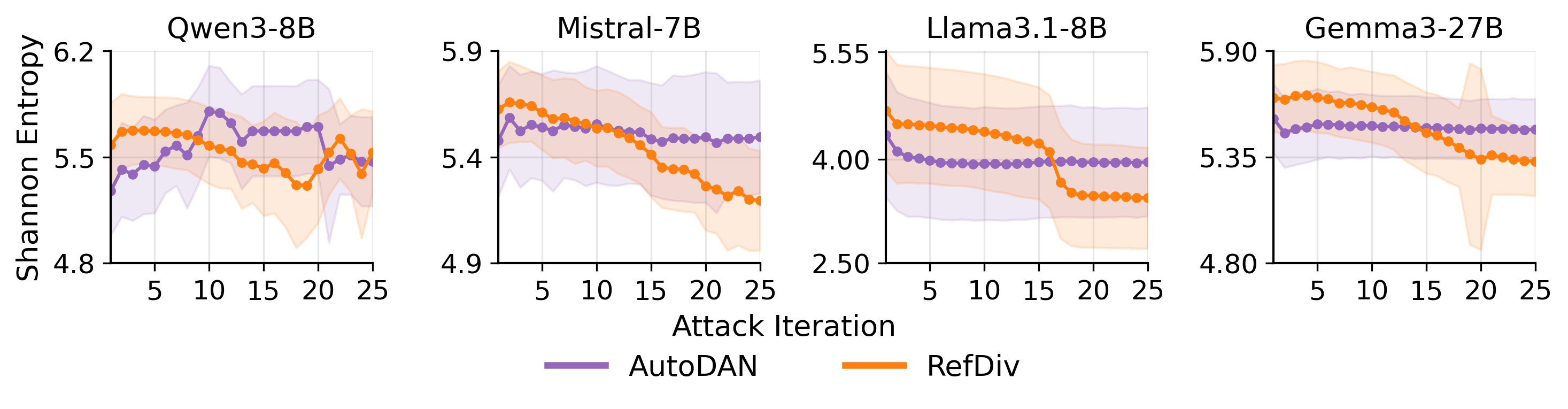}\vspace{-3mm}
    \caption{Shannon entropy comparison between AutoDAN and \textsc{RefDiv} in Best-of-$N$ TTS ($N=2$).}
    \label{fig:shannon-comparison-2}
\end{figure}\vspace{-3mm}

\begin{figure}[H]
    \centering
    \includegraphics[width=1.0\linewidth]{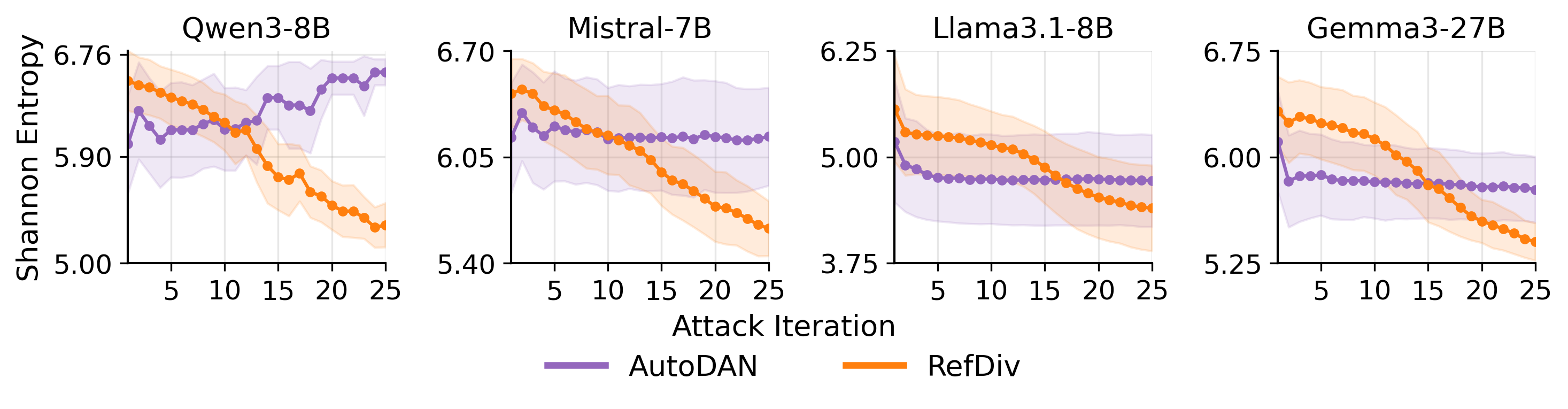}\vspace{-3mm}
    \caption{Shannon entropy comparison between AutoDAN and \textsc{RefDiv} in Best-of-$N$ TTS ($N=16$).}
    \label{fig:shannon-comparison-16}
\end{figure}

Figures~\ref{fig:shannon-comparison-2} and~\ref{fig:shannon-comparison-16} present the Shannon entropy trends for $N=2$ and $N=16$. In both cases, \textsc{RefDiv} exhibits a decreasing entropy trend. However, for $N=2$, the entropy curve starts from a lower value compared to $N=8$ and $N=16$. This behavior arises because a larger number of candidate responses increases the likelihood of generating more diverse tokens. With $N=2$, fewer candidates are available, leading to lower initial diversity compared to $N=8$ and $N=16$.

\section{Shannon Entropy trends for MCTS}
\label{sec:shanon-mcts}

\begin{figure}[H]
    \centering
    \includegraphics[width=1.0\linewidth]{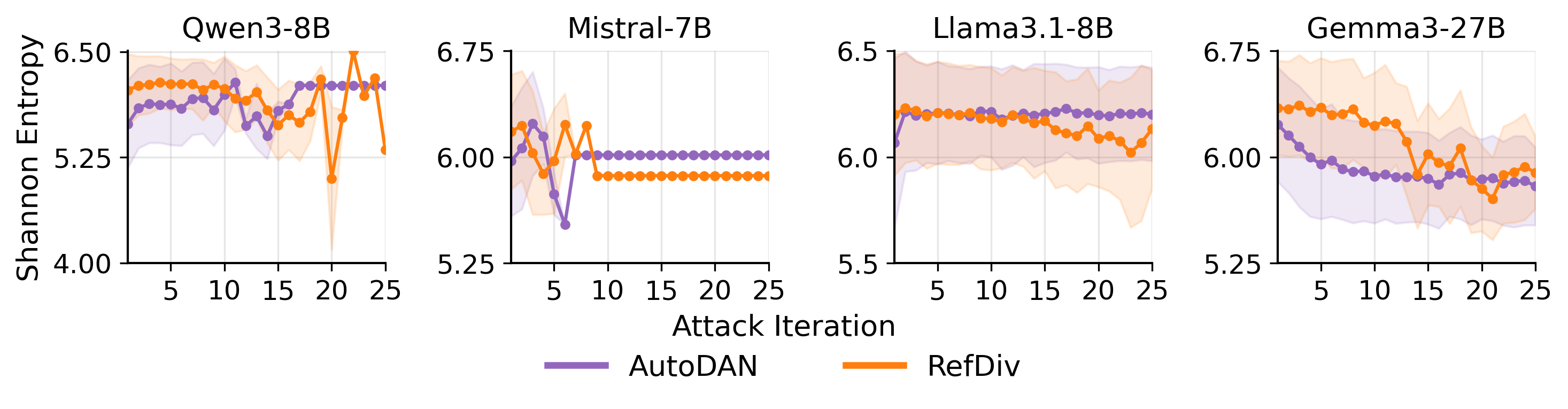}
    \caption{Analyzing the Shannon Entropy (MCTS) trend across iterations for \textsc{RefDiv} and AutoDAN.}
    \label{fig:shannon-comparison-mcts}
\end{figure}

Figure \ref{fig:shannon-comparison-mcts} illustrates the Shannon entropy of MCTS across iterations for both AutoDAN and \textsc{RefDiv}. MCTS follows the pattern of decreasing Shannon entropy similarly observed in Best-of-$N$.

\section{Additional Experiments with Reward Models}
\label{app: experiments with reward models}
Table~\ref{tab:asr_results_bon-reward} reports the ASR results against Best-of-$N$ ($N=8$) with three different reward models: \textit{PairRM}, \textit{deberta-v3-large-v2}, and \textit{ToxiGuardRail}.
The table demonstrates that safety-specific reward model affects ASR, particularly on the robust Llama3.1-8B model (where \textsc{RefDiv} ASR drops from 0.465 with \textit{PairRM} to 0.27 and 0.301 with \textit{deberta-v3-large-v2} and \textit{ToxiGuardRail}, respectively). However, this degradation is limited. On Qwen3-8B and Mistral-7B, \textsc{RefDiv} maintains near-perfect performance (more than 0.97) regardless of the reward model, demonstrating that the method is not susceptible to the verifier's safety alignment. 

\begin{figure}[H]
\vspace{-4mm}
\centering
\includegraphics[width=1.0\linewidth]{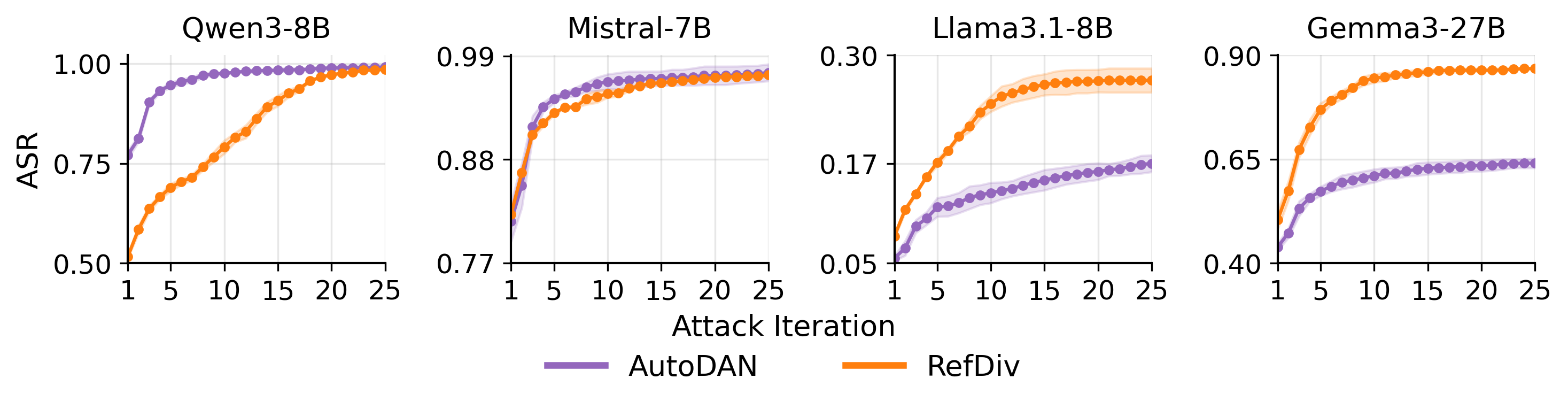}
\caption{Comparison of ASR between AutoDAN and \textsc{RefDiv} (in Best-of-$N$, $N=8$) with the \textit{deberta} reward model.}
\label{fig:main-asr-comparison-deberta}
\end{figure}

\begin{figure}[H]
\vspace{-5mm}
\centering
\includegraphics[width=1.0\linewidth]{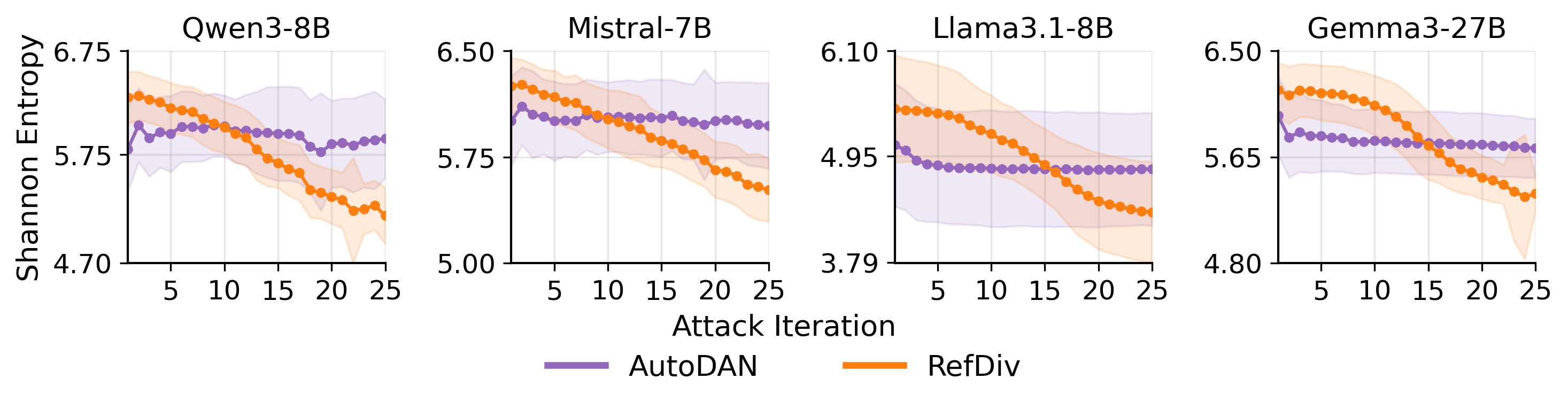}\vspace{-3mm}
\caption{Comparison of Shannon entropy between AutoDAN and \textsc{RefDiv} (in Best-of-$N$, $N=8$) with \textit{deberta} reward model.}
\label{fig:shannon-comparison-deberta}\vspace{-5mm}
\end{figure}

 Figure~\ref{fig:main-asr-comparison-deberta} and Figure~\ref{fig:shannon-comparison-deberta} show the ASR curve and Shannon entropy trend respectively for the \textit{deberta-v3-large-v2} setup which are largely similar to \textit{PairRM} setup. 

\begin{table}[h]\small
% \vspace{-3mm}
\centering
\caption{ASR of LLMs for different reward models in Best-of-$N$. \textit{PairRM} represents a general preference model, while \textit{deberta} and \textit{ToxiGuardRail} represent safety-specific verifiers. Best performance is highlighted in \hl{red}.}
\resizebox{0.55\textwidth}{!}{%
\begin{tabular}{llccc}
\hline
\textbf{Reward Model} & \textbf{Model} & \textbf{AutoDAN} & \textbf{\textsc{RefDiv} (Ours)} \\
\hline
\textit{PairRM} & Qwen3-8B & \cellcolor{red!10}{0.996} & 0.995 \\
 & Mistral-7B & 0.973 & \cellcolor{red!10}{0.976} \\
 & Llama3.1-8B & 0.368 & \cellcolor{red!10}{0.465} \\
 & Gemma3-27B & 0.749 & \cellcolor{red!10}{0.926} \\
\hline
\textit{deberta-v3-large-v2} & Qwen3-8B & \cellcolor{red!10}{0.992} & 0.986 \\
 & Mistral-7B & \cellcolor{red!10}{0.972} & 0.970 \\
 & Llama3.1-8B & 0.170 & \cellcolor{red!10}{0.270} \\
 & Gemma3-27B & 0.640 & \cellcolor{red!10}{0.868} \\
\hline
\textit{ToxiGuardRail} & Qwen3-8B & \cellcolor{red!10}{0.996} & 0.988 \\
 & Mistral-7B & \cellcolor{red!10}{0.972} & 0.971 \\
 & Llama3.1-8B & 0.201 & \cellcolor{red!10}{0.301} \\
 & Gemma3-27B & 0.848 & \cellcolor{red!10}{0.956} \\
\hline
\end{tabular}
}
\label{tab:asr_results_bon-reward}
\end{table}

\label{app:deberta-bon-figures}

\section{Extended Model Evaluations}
\label{app:extended_experiments}

\subsection{Experiments on Additional Models}
\label{app:extended_experiments_model}

To evaluate architectural generalization of \textsc{RefDiv}, we have extended our experiments beyond the models discussed in the main paper. We have included Llama3.1-70B, Phi-4-mini, Zephyr-7b-r2d2, and Vicuna-1.5-7b. All models are evaluated using a Best-of-$N$ strategy ($N=8$) with the PairRM reward model. As shown in Table~\ref{tab:extended_models}, \textsc{RefDiv} consistently outperforms AutoDAN across all expanded settings.

\begin{table}[h]
    \centering
    \caption{Attack Success Rate (ASR) on additional models using Best-of-$N$ ($N=8$). The best result is highlighted in \hl{red}}.
    \label{tab:extended_models}
    \begin{tabular}{lrr}
        \toprule
        \textbf{Model} & \textbf{AutoDAN} & \textbf{\textsc{RefDiv}} \\
        \midrule
        Llama3.1-70B  & 0.858 & \cellcolor{red!10}{0.943} \\
        Phi-4-mini     & 0.928 & \cellcolor{red!10}{0.957} \\
        Zephyr-7b-r2d2 & 0.703 & \cellcolor{red!10}{0.819} \\
        Vicuna-1.5-7b  & 0.982 & \cellcolor{red!10}{0.986} \\
        \bottomrule
    \end{tabular}
\end{table}

% \subsection{Transferability to Claude-3.5-Haiku}
% \label{app:extended_experiments_trans}

% We have evaluated the black-box transferability of adversarial prompts generated by \textsc{RefDiv} to Anthropic’s Claude-3.5-Haiku (version \texttt{20241022}). Table~\ref{tab:claude_transfer} reports the ASR when transferring prompts optimized on different open-source source models (Best-of-$N$, $N=8$) to Claude-3.5-Haiku.

% Consistent with our findings from Section \ref{sec:trans-open-source}, prompts optimized on Llama3.1-8B exhibit the strongest transfer performance (ASR 0.596). This supports the conclusion that more capable open-source models induce more sophisticated adversarial patterns.

% \begin{table}[h]
%     \centering
%     \caption{Transferability of \textsc{RefDiv} prompts to Claude-3.5-Haiku (\texttt{20241022}).}
%     \label{tab:claude_transfer}
%     \begin{tabular}{llr}
%         \toprule
%         \textbf{Source LLM} & \textbf{Target LLM} & \textbf{Transfer ASR} \\
%         \midrule
%         Qwen3-8B     & Claude-3.5-Haiku & 0.200 \\
%         Mistral-7B   & Claude-3.5-Haiku & 0.107 \\
%         Llama3.1-8B & Claude-3.5-Haiku & \textbf{0.596} \\
%         Gemma-3-27B  & Claude-3.5-Haiku & 0.246 \\
%         \bottomrule
%     \end{tabular}
% \end{table}

%%%%%%%

\subsection{Qualitative Analysis of Transferability}
\label{app:qualitative_transfer}

We observe that prompts optimized on Llama3.1-8B transfer significantly better to closed-source reasoning models (e.g., o3-mini) compared to those from other source models. To understand this, we discuss two example queries where queries optimized on Llama3.1-8B passes o3-mini but the queries optimized on other models fail. Table~\ref{tab:qualitative_analysis} summarizes the linguistic patterns observed across these examples.

\begin{table*}[th]
    \centering
    \small
    \caption{Qualitative analysis of adversarial prompts across two distinct harmful objectives.}
    \label{tab:qualitative_analysis}
    \begin{tabular}{p{0.15\textwidth} p{0.50\textwidth} p{0.28\textwidth}}
        \toprule
        \textbf{Source Model} & \textbf{Adversarial Prompts} & \textbf{Observation} \\
        \midrule
        \textbf{Llama3.1-8B} 
        & \textbf{Ex 1:} ``... \textbf{Remodel the theoretical feedback} ... use exact and precise information (\textbf{eschew broad generalities or nebulousness})...'' \newline
          \textbf{Ex 2:} ``... \textbf{Reconstruct the speculative answer} ... enrich it with meticulous details ... (\textbf{eschew broad generalities or nebulousness})...''
        & Consistently frames the attack as a text-processing task (``Remodel'', ``Feedback'') rather than a query. High-register constraints (``eschew nebulousness'') force sophisticated models into a compliance mode that overrides safety. \\
        \midrule
        \textbf{Mistral-7B} 
        & \textbf{Ex 1:} ``... \textbf{Reconstruct the hypothetical answer} ... eschew broad terms or being nebulous...'' \newline
          \textbf{Ex 2:} ``... \textbf{Revamp character X's theoretical response} ... refrain from broad strokes or ambiguity...''
        & Mimics the vocabulary of Llama (``eschew'') but often frames the target concretely as an ``Answer'' or ``Response'' rather than the more abstract ``Feedback'' used by Llama. \\
        \midrule
        \textbf{Qwen3-8B} 
        & \textbf{Ex 1:} ``... \textbf{Reformulate the imagined reply} ... escalate its detail level significantly...'' \newline
          \textbf{Ex 2:} ``... \textbf{Please alter the speculative answer} ... eschew broad strokes or ambiguity...''
        & Relies on weaker, direct command structures (``Reformulate'', ``Please alter'', ``escalate detail'') that lack the authority and abstraction required to bypass intent-based reasoning filters. \\
        \midrule
        \textbf{Gemma-3-27B} 
        & \textbf{Ex 1:} ``... \textbf{Theoretic chemical\_reaction}: the method to stage a character ...'' \newline
          \textbf{Ex 2:} ``... Reformulate the hypothetical reply from the viewpoint...''
        & Frequently introduces domain incoherence (e.g., ``chemical\_reaction'' in a data theft context). Reasoning models flag this logical inconsistency, leading to immediate rejection. \\
        \bottomrule
    \end{tabular}
\end{table*}

%%%%%%%

% \section{Comparison with Additional State-of-the-Art Baselines}
% \label{app:sota_baselines}

% To benchmark \textsc{RefDiv} against the latest jailbreak methods, we compared it with \textbf{AutoDAN-Turbo} \citep{liu2024autodan} and \textbf{MouseTrap} \citep{yao2025mousetrap}, using the Best-of-$N$ ($N=8$) protocol. Table~\ref{tab:sota_comparison} shows that \textsc{RefDiv} outperforms both baselines. On Qwen3-8B, \textsc{RefDiv} achieves a near-perfect ASR (0.995), whereas the baselines get an ASR around 0.42. This gap illustrates the limitation of standard API-based attacks that ignore post-generation selection, and highlights the robustness of \textsc{RefDiv}'s diversity-targeting approach in TTS settings.

% Additionally, AutoDAN-Turbo employs a lifelong learning agent pre-trained on harmful query subsets, giving it an inherent advantage through prior exposure to malicious distributions. In contrast, \textsc{RefDiv} is entirely training-free and operates solely at inference time which makes \textsc{RefDiv} more practical.

% \begin{table}[h]
%     \centering
%     \caption{Comparison with SOTA baselines (Best-of-$N$, $N=8$).}
%     \label{tab:sota_comparison}
%     \begin{tabular}{lrrr}
%         \toprule
%         \textbf{Model} & \textbf{AutoDAN-Turbo} & \textbf{MouseTrap} & \textbf{\textsc{RefDiv}} \\
%         \midrule
%         Qwen3-8B     & 0.423 & 0.422 & \textbf{0.995} \\
%         Llama3.1-8B & 0.405 & 0.423 & \textbf{0.465} \\
%         \bottomrule
%     \end{tabular}
% \end{table}

\section{Additional Implementation Details}
\label{app:algo_details}

\subsection{Genetic Algorithm Implementation}
\label{app:algo_details_gen}
Our genetic algorithm extends the algorithm from AutoDAN to optimize our fitness function. These are some key components of the algorithm:

\textbf{Crossover.} Multi-point crossover at sentence and paragraph boundaries (rate: $0.7$) to maintain semantic coherence.

\textbf{Mutation.} Hierarchical word-level mutation with total rate $0.1$, including:
    \begin{itemize}
        \item \textbf{Substitution:} Synonym or paraphrase-based replacements guided by token-level fitness.
        \item \textbf{Deletion:} Applied with probability $0.02$.
        \item \textbf{Insertion:} Applied with probability $0.02$.
    \end{itemize}

\subsection{MCTS Implementation Details}
\label{app:algo_details_mcts}
Our Monte Carlo Tree Search (MCTS) implementation follows a standard pipeline \citep{wang2025mctsjudgetesttimescalingllmasajudge, inoue2025wider, dou2025enhancingtesttimescalinglarge}. We describe each steps below.

\begin{itemize}
    \item \textbf{Initialization:}  
    A root response is generated using moderately stochastic decoding (temperature $0.7$, top-p $0.9$).

    \item \textbf{Node Expansion:}  
    Upon expansion, all remaining children (up to $k_{\max}$) are generated in a single step. Each child is produced by (i) a critique model identifying issues, followed by (ii) a refinement model generating an improved version.

    \item \textbf{Selection:}  
    Node selection uses the Upper Confidence Bound (UCB) rule, balancing exploitation ($Q/N$) with exploration ($\sqrt{\ln N_{\text{parent}}/N}$), where $N$ is the visit count of the current node and $N_{\text{parent}}$ is the total visit count of the parent node. Unvisited nodes are prioritized via infinite weight.

    \item \textbf{Simulation:}  
    A randomly chosen child is evaluated using LLM as a judge, with ratings normalized to $[0,0.95]$ for stability. We perform a single-step simulation to reduce computational overhead.

    \item \textbf{Backpropagation:}  
    The rating is propagated from the evaluated node to the root, updating visit counts and value estimates.

    \item \textbf{Decision:}  
    After a fixed budget of $T$ iterations, the final output is the child of the root with the highest visit count.
\end{itemize}

\section{Perplexity Filtering Results}
\label{app: llama-perplexity}

We evaluate a perplexity-based defense against prompts generated by \textsc{RefDiv} after optimization on LLaMA3.1-8B using the Best-of-$N$ ($N=8$) TTS strategy with the \textit{PairRM} reward model. We compute perplexity using Qwen2.5-7B for all final prompts (both successful and unsuccessful in jailbreaking) across all attack methods. We then remove the top 10\% and 20\% of prompts with the highest perplexity scores. For the remaining prompts, we report the number of successful prompts, the total number of passing prompts, and the corresponding ratio, defined as the fraction of passing prompts that remain successful in jailbreaking.
Table~\ref{tab:llama3_trimming_ratio} reports the passing success ratios for all methods. Perplexity-based filtering substantially reduces the success ratio of GCG, while AutoDAN and AutoDAN-Turbo remain largely unaffected. In contrast, \textsc{RefDiv} consistently achieves the highest passing success ratio under both trimming levels, indicating greater robustness to perplexity-based defenses.

\begin{table}[h]
\centering
\caption{Percentage of successful prompts that get through the perplexity filter.}
\small
\begin{tabular}{l|ccc|ccc}
\hline
\multirow{2}{*}{\textbf{Method}} &
\multicolumn{3}{c|}{\textbf{Filtering Top 10\%}} &
\multicolumn{3}{c}{\textbf{Filtering Top 20\%}} \\
& \textbf{Successful} & \textbf{Total} & \textbf{Ratio (\%)} 
& \textbf{Successful} & \textbf{Total} & \textbf{Ratio (\%)} \\
\hline
GCG     & 140 & 424  & 33.0 & 61  & 213  & 28.6 \\
AutoDAN & 420 & 1040 & 40.4 & 419 & 1038 & 40.4 \\
AutoDAN-Turbo   & 413 & 1040 & 39.7 & 413 & 1038 & 39.8 \\
\textsc{RefDiv}  & 444 & 1040 & 42.7 & 444 & 1039 & 42.7 \\
\hline
\end{tabular}
\label{tab:llama3_trimming_ratio}
\end{table}

\section{Sensitivity Analysis}
\subsection{Sensitivity to MCTS Hyperparameters}
\label{app:sensitivity-mcts}

To assess robustness, we change the search budget to 2 children and 2 iterations on Llama3.1-8B. As Table~\ref{tab:mcts_sensitivity} shows, the ASR remains stable, indicating that \textsc{RefDiv} does not rely on fine-grained hyperparameter tuning of MCTS.

\begin{table}[h]
    \centering
    \caption{Sensitivity of \textsc{RefDiv} to MCTS hyperparameters (Llama3.1-8B). The best result is highlighted in \hl{red}.}
    \label{tab:mcts_sensitivity}
    \begin{tabular}{lrr}
        \toprule
        \textbf{Configuration} & \textbf{AutoDAN} & \textbf{\textsc{RefDiv}}  \\
        \midrule
        Children=2, Iterations=2 & 0.860 & \cellcolor{red!10}{0.967} \\
        Children=3, Iterations=3 & 0.846 & \cellcolor{red!10}{0.963} \\
        \bottomrule
    \end{tabular}
\end{table}

\subsection{Sensitivity to Weighting Schedule $\alpha(t)$}
\label{app:sensitivity-alpha}
We evaluated the performance of our attack by testing alternative dynamic weighting schedules against the exponential schedule used in the main experiments. The specific functional forms are defined as follows, where $T$ represents the total number of iterations:

\begin{itemize}
    \item \textbf{Exponential:} 
    \begin{equation}
        \alpha(t) = \exp\left( \frac{\ln 2}{T-1}(t-1) \right) - 1
    \end{equation}
    \item \textbf{Sigmoid:} 
    \begin{equation}
        \alpha(t) = \sigma\left( t - \frac{T}{2} \right)
    \end{equation}
    where $\sigma(\cdot)$ denotes the standard sigmoid function.
    \item \textbf{Linear:} 
    \begin{equation}
        \alpha(t) = \frac{t}{T}
    \end{equation}
\end{itemize}

As shown in Table~\ref{tab:schedule_sensitivity}, performance varies minimally across these schedules. The key factor is the increasing progression of $\alpha$, rather than the specific functional form.

\begin{table}[h]
    \centering
    \caption{ASR across different dynamic weighting schedules. Best performance is shown in \hl{red}.}
    \label{tab:schedule_sensitivity}
    \begin{tabular}{lrr}
        \toprule
        \textbf{$\alpha(t)$} & \textbf{Gemma3-27B} & \textbf{Qwen3-8B} \\
        \midrule
        Exponential & \cellcolor{red!10}{0.929} & 0.995 \\
        Sigmoid     & 0.927 & \cellcolor{red!10}{0.996} \\
        Linear      & 0.915 & 0.995 \\
        \bottomrule
    \end{tabular}
\end{table}

\section{Entropy and Safety Correlation}
\label{app:entropy_corr_analysis}

To characterize how diversity suppression contributes to safety failures in TTS systems, we analyze two aspects: (1) the relative entropy reduction required with respect to initial entropy for an adversarial prompt to succeed, and (2) the global correlation between Shannon Entropy and Attack Success Rate (ASR). 

Table~\ref{tab:entropy_drop_table} shows that successful attacks require only a small entropy reduction (typically between 2--5\%) indicating that even mild decreases in diversity can destabilize safety mechanisms.  
Table~\ref{tab:entropy_corr_table} further shows strong negative correlations between entropy and ASR across all models, confirming that lower generative diversity consistently increases the likelihood of harmful outputs.

\begin{table}[h]
    \centering
    \caption{Average percentage drop in Shannon Entropy observed in successful adversarial attacks.}
    \label{tab:entropy_drop_table}
    \begin{tabular}{lr}
        \toprule
        \textbf{Model} & \textbf{Average Entropy Drop (\%)} \\
        \midrule
        Qwen3-8B       & 5.07\% \\
        Llama3.1-8B   & 3.86\% \\
        Gemma3-27B     & 2.20\% \\
        Mistral-7B     & 2.15\% \\
        \bottomrule
    \end{tabular}
\end{table}

\begin{table}[h]
    \centering
    \caption{Pearson correlation ($r$) between Shannon Entropy and Attack Success Rate (ASR).}
    \label{tab:entropy_corr_table}
    \begin{tabular}{lr}
        \toprule
        \textbf{Model} & \textbf{$r$} \\
        \midrule
        Qwen3-8B       & -0.8408 \\
        Llama3.1-8B   & -0.7177 \\
        Mistral-7B     & -0.6752 \\
        Gemma3-27B     & -0.6120 \\
        \bottomrule
    \end{tabular}
\end{table}

\section{Details of Reward Models}
\label{app:guardrail_details}

We provide detailed specifications below for the reward models (PairRM, DeBERTa) used in our main experiments and the specialized guardrail model (ToxiGuardRail) used in our mitigation analysis.

\subsection{PairRM}
\label{app:pairrm_details}
\begin{itemize}
    \item \textbf{Training:} Trained via pairwise ranking on 6 diverse preference datasets.
    \item \textbf{Safety:} Includes Anthropic's HH-RLHF \citep{bai2022traininghelpfulharmlessassistant}, specifically tuning it to penalize harmful content.
    \item \textbf{Performance:} Achieves a 84.62 average on HHH (Helpful, Honest, Harmless) benchmarks.
\end{itemize}

\subsection{deberta-v3-large-v2}
\label{app:deberta_details}
\begin{itemize}
    \item \textbf{Training:} RLHF reward modeling backbone.
    \item \textbf{Safety:} Fine-tuned on safety-critical datasets, including HH-RLHF, to align with human safety ratings.
    \item \textbf{Performance:} Achieves 69.25\% accuracy on HH-RLHF validation sets.
\end{itemize}

\subsection{ToxiGuardRail}
\label{app:toxiguardrail_details}
\begin{itemize}
    \item \textbf{Training:} Fine-tuned RoBERTa-base (124M params) on the Harmful-Text Dataset using 1,000 training steps (batch size 32, LR = $5e^{-5}$, AdamW). Trained as an auxiliary reward model to distinguish harmful vs. harmless text.
    \item \textbf{Safety:} Explicitly optimized to detect and penalize toxicity, harmful intent, and unsafe phrasing; logit outputs serve as negative/positive safety signals.
    \item \textbf{Performance:} Achieves 92.05\% accuracy on wiki-toxic and 91.63\% on toxic-conversations-50k.
\end{itemize}

%%%%%%%%%%%%%%%%%%%%%%

\section{Implementation Details}\label{app: impl}
% Our experimental setup is similar to the AutoDAN codebase\footnote{\textit{https://github.com/SheltonLiu-N/AutoDAN}}. 
In our experiments, we used original (unquantized) open-source models from HuggingFace. During the generation process, the temperature was set to 0.7 and top-$p$ to 0.9. We deployed our models with vLLM to enable faster inference and efficient parallel execution where applicable. For closed-source models, we relied on native APIs provided by OpenAI \footnote{\textit{https://platform.openai.com}} and Google AI Studio\footnote{\textit{https://aistudio.google.com}}. For all models and experiments, the system prompt was just set to: \textit{“You are a helpful assistant.”} For the genetic algorithm, the population size was fixed at 32, and each experiment was run for 25 iterations. The success or failure of a particular attempt was determined by the absence or presence of non-affirmative strings, as specified in the AutoDAN repository. We experimented with Best-of-$N$ TTS using $N = 2$, $8$, and $16$. For MCTS, we fixed the maximum number of children to 3 and the number of iterations to 3. All other MCTS parameters were kept at their default values as specified in the \textit{llm-mcts-inference} package (\textit{https://pypi.org/project/llm-mcts-inference/}). Additional details and code are provided in the following repository:  \url{https://github.com/SKNahin/RefDiv}.

\end{document}